\documentclass[lettersize,journal]{IEEEtran}
\usepackage{amsmath,amsfonts}
\usepackage{algorithmic}
\usepackage{algorithm}
\usepackage{array}
\usepackage[caption=false,font=normalsize,labelfont=sf,textfont=sf]{subfig}
\usepackage{textcomp}
\usepackage{stfloats}
\usepackage{url}
\usepackage{verbatim}
\usepackage{graphicx}
\usepackage{cite}
\usepackage{cleveref}
\usepackage{xcolor}
\usepackage{booktabs}
\usepackage{multirow}
\usepackage{xcolor}
\usepackage{tikz}
\usepackage{diagbox}
\usetikzlibrary{spy,calc}

\usepackage{bbding}

\newcommand{\mymodel}{SpatioTemporally Aligned Visual Prompt Tuning~}

\newcommand{\pei}[1]{\textcolor[rgb]{0,0,1} {[Pei: #1]}}

\begin{document}
%
\title{STA-VPT: SpatioTemporally Aligned Visual Prompt Tuning}

\author{Wenjie Pei,~\IEEEmembership{Senior~Member,~IEEE,}
        Tongqi Xia,
        Qizhong Tan,
        Jiandong Tian,
        Guangming Lu,~\IEEEmembership{Senior~Member,~IEEE,}
        Jun Yu,~\IEEEmembership{Senior~Member,~IEEE}
\thanks{Wenjie Pei, Qizhong Tan, Guangming Lu and Jun Yu are with Harbin Institute of Technology, Shenzhen, China. Tongqi Xia is with vivo Mobile Communication Co., Ltd. Jiandong Tian is with Shenyang Institute of Automation, Chinese Academy of Sciences. Wenjie Pei is also with Pengcheng Laboratory, Shenzhen, China.}

\thanks{Wenjie Pei and Tongqi Xia contibute equally to this work.}

\thanks{Wenjie Pei is the corresponding author (wenjiecoder@outlook.com).}

}

\markboth{Preprint}{Preprint}


\maketitle

\begin{abstract}
Typical methods for visual prompt tuning follow the sequential modeling paradigm originating from NLP, learning a sequence of unordered parameterized tokens as visual prompts, which are then prefixed to the flattened image representation for model adaptation. While such a sequential prompting paradigm has exhibited great promise, it presents two potential limitations. First, the learned visual prompts, presented in an unordered sequential form, are unable to capture the underlying spatial relations in the input image which are crucial for effective image encoding. Second, all prompt tokens serve the same role by performing uniform prompting for all image tokens without distinction, lacking fine-grained prompting capability\---i.e., individualized prompting for different visual tokens to capture region-specific semantic patterns. In this work, we introduce the \mymodel model (\emph{STA-VPT}), a novel visual prompting paradigm, which learns a two-dimensional prompt token map for image prompting or a three-dimensional token volume for video prompting, ensuring spatial (or spatiotemporal) alignment with the input image token map (or video token volume). This alignment enables the visual prompts to preserve the spatial (or spatiotemporal) structure, thereby learning the underlying relations within the visual input. Furthermore, each prompt token serves as a specialized prompting expert and is designated to exclusively prompt for the spatially (or spatiotemporally) corresponding visual tokens. Consequently, our \emph{STA-VPT} is capable of performing individualized prompting for different spatiotemporal regions, potentially improving the prompting performance through fine-grained allocation of prompting capacity, in line with mixture‑of‑experts (MoE) principles. We demonstrate the superior performance and potential merits of \emph{STA-VPT} on both image prompting and video prompting across diverse downstream tasks, including image classification, semantic segmentation, and action recognition.
\end{abstract}

\begin{IEEEkeywords}
Parameter-efficient fine-tuning, Visual prompt tuning.
\end{IEEEkeywords}

\section{Introduction}
\IEEEPARstart{A}{dapting} large pre-trained vision models to a specific downstream vision task via parameter-efficient fine-tuning (PEFT) has proven to be an effective and efficient strategy, particularly in the scenarios where the training data is limited. PEFT not only distills and leverages the prior knowledge relevant to the downstream task from the pre-trained model to achieve better performance, but also significantly enhances the training efficiency by incurring minimal parameter-tuning overhead. As a prominent PEFT method, prompt tuning has achieved remarkable success in NLP field and is being explored its potential in computer vision~\cite{ref27,ref31,ref28,ref57,han20232vpt,ref29}. 

\begin{figure}[!t]%
\centering
\includegraphics[width=1.0\linewidth]{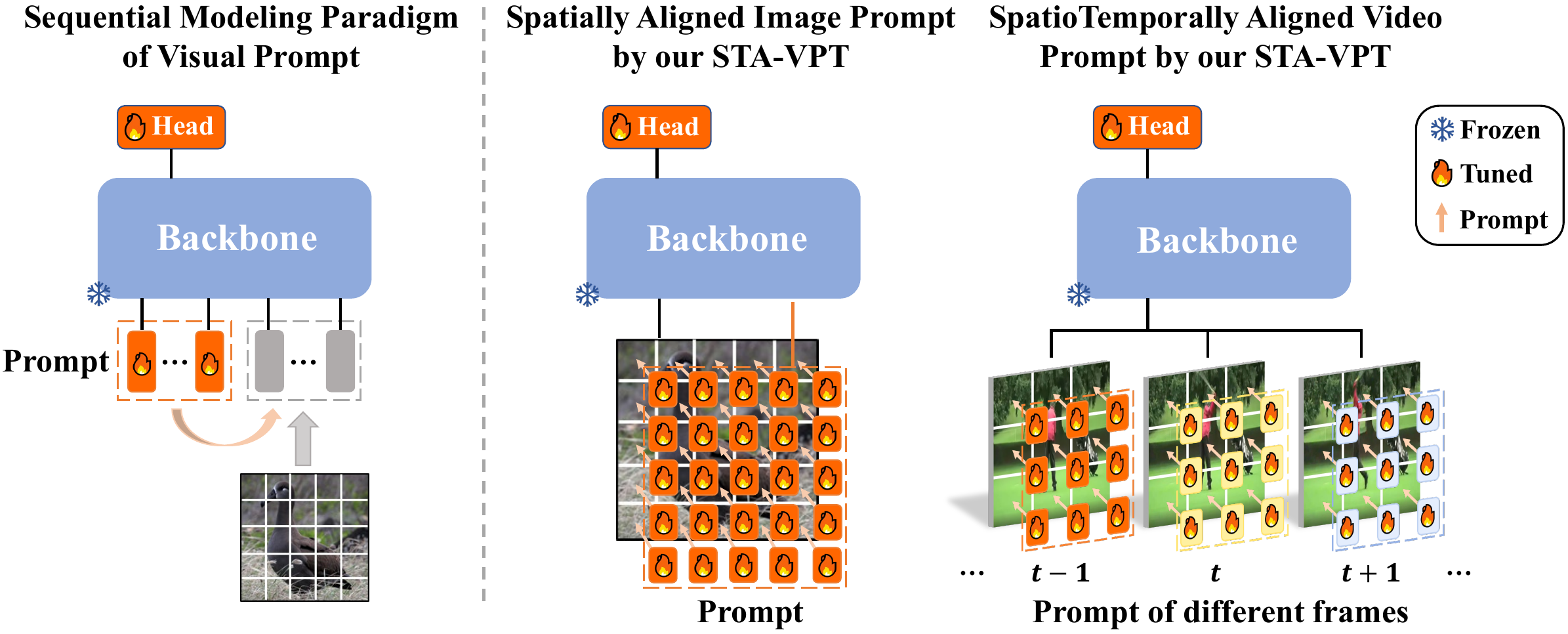}
\caption{Typical methods for visual prompting tuning follow the sequential modeling paradigm which learns an unordered sequence of prompts prefixed to the flattened image representation. All prompt tokens serve the same role in performing uniform prompting for all image tokens. In contrast, our \emph{STA-VPT} learns a 2D prompt token map for image prompting or a 3D token volume for video prompting, which aligns with the image token map spatially (or the video token map spatiotemporally). As a result, our \emph{STA-VPT} is able to 1) conduct individualized prompting for different visual tokens in a fine-grained manner and 2) model the spatial (spatiotemporal) relations in the input image (video) by preserving the spatial (spatiotemporally) structure in the prompts.}
\label{fig:teaser}
\end{figure}%

Most existing methods for visual prompt tuning~\cite{ref28,ref57,han20232vpt} follow the sequential modeling paradigm, which originates from the  prompt learning in NLP field, as shown in Figure~\ref{fig:teaser}. A prominent example is VPT~\cite{ref28}, which encodes an input image as a sequential representation using pre-trained vision models and then learns an unordered sequence of prompt tokens as auxiliary input prefixed to the image representation. The learned vision prompts are expected to distill relevant knowledge from the pre-trained model, thereby adapting the pre-trained model to the downstream task. While such sequential modeling paradigm of visual prompts has demonstrated great promise, it faces two potential limitations. First, the unordered sequential structure of prompt tokens fails to capture the spatial structure and model the underlying spatial relations of the input image, which is essential for visual feature learning. Second, all visual prompts in the sequential modeling paradigm serve the same role, performing uniform prompting for all image tokens without distinction. As a result, it is unable to conduct fine-grained, individualized prompting for different image tokens. However, image tokens in different regions typically exhibit distinct semantic patterns, necessitating region-specific prompting.

To address the aforementioned two limitations, in this work we propose a novel visual prompting paradigm, dubbed the SpatioTemporally Aligned Visual Prompt Tuning model (\emph{STA-VPT}). As illustrated in Figure~\ref{fig:teaser}, instead of learning an unordered sequence of visual prompts in the sequential prompting paradigm, our \emph{STA-VPT} learns a two-dimensional prompt token map for image prompting or a three-dimensional prompt token volume for video prompting, which matches the dimensional shape of the input image token map or video token volume. Besides, the learned positional embeddings of the pre-trained model are incorporated into the prompt token map to preserve the spatial structure. As a result, \emph{STA-VPT} is able to align the prompt token map (or volume) with the visual input spatially (or spatiotemporally). 

By establishing the spatiotemporal alignment between prompt and visual tokens, \emph{STA-VPT} 
surmounts the two fundamental limitations of the sequential prompt paradigm. 
First, Such alignment enables the visual prompts to learn the underlying spatial (or spatiotemporal) relations within the input image (or video), as the spatial (spatiotemporal) structure is well preserved in the prompt tokens. This allows our model to potentially distill more effective knowledge through the visual prompts. Second, each prompt token is designated to exclusively prompt for the spatially (or spatiotemporally) corresponding visual tokens, enabling fine-grained, individualized prompting mechanism and highly granular adaptation to diverse visual patterns. In contrast to the uniform prompting by all prompt tokens in the sequential prompt paradigm, each prompt token in our \emph{STA-VPT} acts as a specialized expert, distilling and prompting knowledge for its spatiotemporally aligned visual tokens. In this sense, \emph{STA-VPT} aligns with the core principles of Mixture of Experts (MoE) models, enhancing performance through a fine-grained allocation of learning capacity that facilitates task-specific expertise. 

To conclude, we make following contributions:
\begin{itemize}
    \item We propose \emph{STA-VPT}, a novel visual prompting paradigm, whose prompting token structure aligns with the dimensional shape of the visual input-either in the spatial domain for image prompting or in the spatiotemporal domain for video prompting. Each prompt token serves as a specialized expert focusing on prompting exclusively for the spatiotemporally aligned visual tokens. This design enables our proposed \emph{STA-VPT} to not only model the underlying spatial (or spatiotemporal) relations in images (or videos), but also perform fine-grained, individualized prompting to capture region-specific semantics.
    \item We develop an effective yet efficient technical implementation for our \emph{STA-VPT}, seamlessly supporting both image prompting and video prompting. It adopts a siamese dual-pathway architecture, equipped with bilateral interaction between two pathways through a novel spatiotemporally aligned cross attention mechanism specially designed for the framework. 
    \item To comprehensively evaluate the proposed \emph{STA-VPT}, we first conduct experiments on image prompting w.r.t. two typical downstream tasks across multiple challenging benchmarks: image classification and semantic segmentation, corresponding to image-level and pixel-level prediction tasks, respectively. Next, we assess the performance of video prompting of \emph{STA-VPT} on action recognition. These experiments demonstrate that our model compares favorably with other state-of-the-art methods for visual prompting tuning, as well as other PEFT methods. Additionally, extensive ablation studies are conducted to validate the effectiveness of each essential design of \emph{STA-VPT}.
\end{itemize}

This paper is an extended version of an earlier conference paper~\cite{conf_version}. Compared with the prior version, this longer article features improvements in four key aspects. 
\begin{itemize}
    \item Our method was initially proposed for individualized image prompting with spatial alignment between prompt and visual tokens in the conference version. Following the proposed visual prompting paradigm, we extend our method technically to the video modality, enabling fine-grained video prompting with spatiotemporal alignment.
    \item In the conference paper, we employed image classification as the downstream task to evaluate our method. To provide a more comprehensive evaluation, we further apply our method to semantic segmentation to validate its effectiveness on pixel-level prediction tasks.
    \item More comprehensive explanations and analyses on the rationale and novelty of the proposed \emph{STA-VPT} are provided in `Introduction' (Section~\ref{sec:intro}) and `Approach' (Section~\ref{sec:method}). Besides, we further improve the technical implementation of the initial conference version in several aspects, including the integration of dimension-wise feature scaling into the prompt adapter, optimization of model training protocols, and finer hyper-parameter tuning.
    \item We conduct substantial additional experiments, including both quantitative and qualitative evaluations on semantic segmentation for image prompting and action recognition for video prompting. 
\end{itemize}
\label{sec:intro}

\section{Related Works}
\subsection{Parameter-Efficient Fine-Tuning (PEFT)}
Parameter-efficient fine-tuning (PEFT) methods perform task adaptation by introducing a small amount of additional learnable parameters while freezing the pre-trained model, which is more efficient than the classical full fine-tuning paradigm. PEFT methods typically fall into four categories: prompt tuning (discussed later), partial fine-funing, adapter tuning and reparameterization. 

\noindent\textbf{Partial fine-tuning} methods selectively fine-tune partial parameters of a pretrained model while keeping the remainder frozen, enabling efficient model adaptation. A straightforward way is to fine-tune only the linear classification head dubbed "Head Fine-Tune", which is highly efficient but offers limited adaptation in complex scenarios. In contrast, BIAS~\cite{zaken2022bitfit} adopts a simple yet effective strategy that tunes only the bias terms of the model. In addition, GPS~\cite{zhang2024gradient} proposes a gradient-based importance estimation scheme to identify those task-sensitive parameters for fine-tuning. Similarly, TR-PTS~\cite{luo2025tr} performs Fisher-guided parameter selection to identify most informative parameters for fine-tuning.

\smallskip\noindent\textbf{Adapter tuning}~\cite{ref16, ref17, ref18, ref19, ref20, steitz2024adapters} generally inserts lightweight transformation module into the intermediate layers of large pre-trained models, transforming features for task adaptation while incurring low parameter-learning overhead. A prominent example is AdaptFormer~\cite{ref21}, which designs a plug-and-play module named AdaptMLP to replace the primitive MLP block in ViT. We draw inspiration from Adapter and devise the prompt adapter in our \emph{STA-VPT} to adapt the prompted knowledge to the downstream task. Furthermore, Adapter+~\cite{steitz2024adapters} conducts systematic exploration of adapter placements, internal architectures, and initialization schemes to optimize adapter configurations.

\smallskip\noindent\textbf{Reparameterization} techniques restructure backbone parameters by integrating learnable components without introducing additional inference costs. LoRA~\cite{hu2022lora}, a representative reparameterization approach, employs low-rank matrix decomposition by multiplying two trainable low-rank matrices and additively merging their product into reparameterizable layer weights, such as those in linear transformations. SSF~\cite{lian2022scaling} adapts pre-trained representations to downstream tasks by introducing learnable scaling and shifting operations to features within frozen backbone networks. Besides, SPT~\cite{he2023sensitivity} proposes a sensitivity-aware adaptation framework, which estimates the sensitivity of backbone weights to the target task and then selectively fine-tunes the most sensitive parameters using either adapter tuning or LoRA.


\smallskip\noindent\textbf{\emph{PEFT} for Video Modality.} PEFT for video inherently requires robust modeling of both spatial appearance and temporal dynamics, particularly for tasks such as action recognition demanding strong temporal reasoning. Extensive work has prioritized adapter tuning for image‑to‑video adaptation. A straightforward approach is to employ two vanilla image adapters to adapt the pre-learned spatial and temporal features separately, either in cascaded manner (AIM~\cite{yang2023aim}) or parallel manner (DUALPATH~\cite{dualpath}). ST-Adapter\cite{pan2022st} pioneers a video‑oriented adapter by incorporating 3D Convolution layer into the vanilla adapter to learn the spatiotemporal feature. 

\subsection{Prompt Tuning}
\noindent\textbf{Prompt Tuning in NLP.} Prompt tuning is initially investigated in NLP and aims to guide the pre-trained large language models to extract the relevant knowledge to the downstream task for task adaptation. The prompts can be hand-crafted~\cite{ref22, ref23}, whose performance relies heavily on the human experience, or automatically generated either in the textual form~\cite{shin2020autoprompt,wallace2019universal} or in the parametric modeling way~\cite{ref24,ref25,lester2021power}. A typical way of parameterizing prompts in NLP is to learn  a sequence of learnable tokens as the prompts concatenated to the input embeddings. This sequential prompting paradigm inherently aligns with the language's sequential nature, delivering excellent prompting performance across diverse NLP tasks. 

\smallskip\noindent\textbf{Visual Prompt Tuning.} Inspired by the great success of prompting tuning in NLP, it is being explored extensively in computer vision. It is infeasible to handcraft the visual prompt directly, thus the visual prompts is either represented as some forms of prior information like object bounding box, mask, or salient points within the target objects~\cite{ref29, kirillov2023segment} or parameterized as learnable embeddings~\cite{ref28, ref57, han20232vpt, ref31, ref27, huang2025cvpt}. As a representative parameterized visual prompting method, VPT~\cite{ref28} follows the sequential prompting paradigm from NLP, learning an unordered sequence of visual prompts that are prepended to the flattened image representation. 

Building upon VPT, various refinements have been investigated. SPT-VPT~\cite{wang2024revisiting} introduces an information-aware initialization scheme for prompt tokens, motivated by empirical findings of high mutual information between prompt tokens and image patch tokens. This approach demonstrates superior performance compared to uniform initialization in the VPT. DA-VPT~\cite{ren2025vpt} leverages metric learning to optimize prompt distributions, explicitly learning distance metrics from class-specific semantic data to guide prompt representations. To improve efficiency, $\text{E}^{2}$VPT~\cite{han20232vpt} learns parameterized prompts only for keys and values (`K'/`V') of self-attention, while applying segment‑wise pruning to further reduce computation. `EXPRES'~\cite{ref57} improves on VPT by learning residual prompts that capture prompt variation with depth, rather than introducing independent prompts in deeper layers, thereby yielding progressively evolved prompts in deep layers. CVPT\cite{huang2025cvpt} uses learned prompts as keys and values to perform cross‑attention with image patch tokens, which in this respect is similar to our \emph{STA‑VPT}. Instead of learning visual prompts from scratch, VAPT~\cite{le2026revisit} derives input-adaptive visual prompts from image tokens using token-wise projectors and a feature projector to improve prompt tuning. PRO-VPT~\cite{shang2025pro} adaptively optimizes prompt distribution by iteratively pruning idle prompts and reallocating them to prompt-needed blocks under a nested optimization framework. STPN~\cite{sun2023spatio} presents the first exploration of prompt tuning for video data, incorporating temporal information from neighboring contexts into the sequential visual prompts. 

Collectively, these VPT-based visual prompt models adhere to the sequential prompting paradigm, which is ill suited to visual modalities such as 2D images and 3D videos. 
In contrast, our proposed STA‑VPT is a novel visual prompting paradigm that learns a 2D prompt map for images or a 3D prompt volume for videos, spatially or spatiotemporally aligned with the visual input. Such design can capture the underlying spatial structure in images or spatiotemporal relations in videos while enabling individualized visual prompting and improved adaptation.

\section{Approach}
\label{sec:method}
\subsection{Overview}
Visual prompt tuning aims to adapt a large pre-trained vision model to a downstream task with minimal parameter tuning overhead. It typically learns prompts as auxiliary inputs to guide the model in distilling the relevant knowledge from a pre-trained model for a specific downstream task, thereby enabling the efficient adaptation of the pre-trained model to the task. Formally, given an input image $I$ and a pre-trained vision model $\mathcal{F}_M$, visual prompt tuning can be formulated as:
\begin{equation}
    O = \mathcal{F}_M(P, I).
\label{eqn:overview}
\end{equation}
Here $P$ denotes visual prompts, which can be either hard prompts with predefined content or soft prompts encoded by learnable parameterized embeddings. $O$ is the prediction by $\mathcal{F}_M$ for a downstream task. During prompt tuning, all the parameters in $\mathcal{F}_M$ are frozen while only $P$ in form of soft prompts and potentially a lightweight projection head are tunable, significantly reducing the tuning overhead.

\begin{figure*}[t]%
\centering
\includegraphics[width=0.98\linewidth]{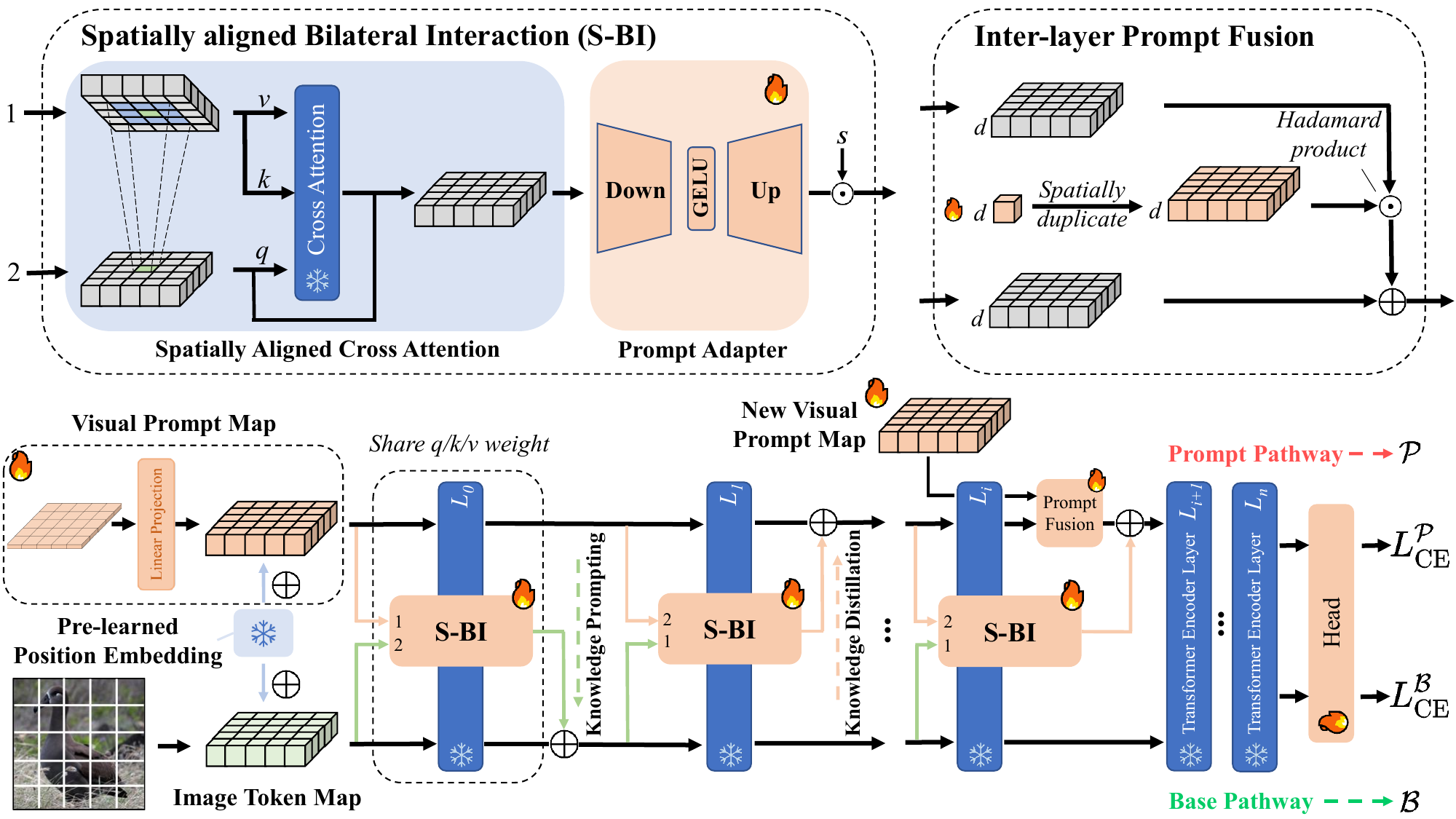}
\caption{Framework of the proposed \emph{STA-VPT} for image prompting with spatial alignment. It follows siamese dual-pathway architecture which consists of Prompt pathway for visual prompt learning and Base pathway for image encoding. Bilateral interaction is performed between two pathways via the designed spatially aligned cross attention. The inter-layer prompt fusion scheme allows for learning multiple prompt maps in different layers.}
\label{fig:framework}
\end{figure*}%

\smallskip\noindent\textbf{Sequential Visual Prompt Tuning Paradigm.} A prominent method for visual prompt tuning is VPT~\cite{ref28}, which follows sequential modeling paradigm originating from the NLP domain. Such sequential prompting paradigm first divides the input image $I$ into $N$ equal-sized patches and encodes $I$ into a flattened sequence of token embeddings $\mathbf{E} \in \mathbb{R}^{N \times d}$, where each of $N$ tokens corresponds to one image patch and $d$ is the feature dimension. Then a set of $c$ parameterized token embeddings $\mathbf{P}\in \mathbb{R}^{c \times d}$ are learned as soft visual prompts and prefixed to the sequence representation $\mathbf{E}$. The concatenated sequence is fed into the pre-trained model $\mathcal{F}_M$ to make predictions for the downstream task. Thus, Equation~\ref{eqn:overview} can be instantiated as:
\begin{equation}
    O = \mathcal{F}_M(\text{Concat}(\mathbf{P}, \mathbf{E})).
    \label{eqn:sequential}
\end{equation}

While the sequential visual prompting paradigm has shown great promise, we investigate two limitations of it: 
\begin{enumerate}
    \item The learned visual prompts in sequential structure cannot capture the spatial relations in the input image which are essential for image encoding.
    \item Since all visual prompt tokens serve the same role in performing uniform prompting for all image tokens, this learning paradigm fails to conduct individualized prompting for different visual tokens, resulting in potential prompting redundancy while restricting the model's prompting efficacy of capturing region-specific semantic patterns.  
\end{enumerate} 

\smallskip\noindent\textbf{SpatioTemporally Aligned Visual Prompt Tuning.} To address aforementioned limitations of the sequential prompting paradigm, we propose \mymodel (\emph{STA-VPT}), a novel visual prompting paradigm, as illustrated in Figure~\ref{fig:framework}. \emph{STA-VPT} learns a visual prompt token map (or volume) $\mathbf{P}^m$ that matches the same dimensional shape as the input image token map (or video token volume) $\mathbf{E}^m$, ensuring spatial (or spatiotemporal) alignment with the input image (or video). This design addresses the key limitations of the sequential prompting paradigm discussed above, yielding two essential benefits. First, it enables the visual prompts to learn spatiotemporal relations within the visual input by preserving the spatiotemporal structure. Furthermore, it allows each prompt token to focus on distilling and prompting knowledge for the spatiotemporally corresponding visual tokens, leading to fine-grained, individualized prompting for different spatiotemporal regions. The prompting mechanism of \emph{STA-VPT} can be expressed as:
\begin{equation}
    O = \mathcal{F}_M(\mathcal{F}_{\text{STA}}(\mathbf{P}^m, \mathbf{E}^m)),
    \label{eqn:sta}
\end{equation}
where $\mathcal{F}_{\text{STA}}$ denotes the spatiotemporally aligned prompting mechanism of the proposed \emph{STA-VPT}. In contrast to the sequential prompt structure learned by VPT ($\mathbf{P}$ in Equation~\ref{eqn:sequential}), the prompt $\mathbf{P}^m$ learned by \emph{STA-VPT} is a two-dimensional token map for image prompting or a three-dimensional token volume for video prompting, ensuring dimensional consistency with the visual input.

We will first explain how the proposed \emph{STA-VPT} performs image prompting with spatial alignment, followed by an elaboration on video prompting with spatiotemporally alignment.

\subsection{Image Prompting with Spatial Alignment}
\label{sec:image prompt}

\subsubsection{Prompt Modeling} We first detail the prompt modeling mechanism ($\mathcal{F}_{\text{STA}}$ in Equation~\ref{eqn:sta}) for image prompting, and then describe the prompt learning for two classical types of downstream vision tasks in the next section: image classification and semantic segmentation.

\smallskip\noindent\textbf{Siamese Dual-Pathway Architecture.} The proposed \emph{STA-VPT} adopts a siamese dual-pathway architecture, as illustrated in Figure~\ref{fig:framework}. Prompt pathway $\mathcal{P}$ is responsible for learning the visual prompts while Base pathway $\mathcal{B}$ focuses on learning visual features based on the prompted knowledge from Prompt pathway. Both pathways employ the same pre-trained vision model with frozen parameters as the backbone. We take ViT-B~\cite{ref2} as an instantiation of the pre-trained vision model while our model is readily applied to other classical vision models like Swin Transformer~\cite{ref4} or ResNet~\cite{he2016deep}. Bilateral interactions between two pathways are performed across the network depth, refining each other iteratively. 

\smallskip\noindent\textbf{Spatially Aligned Prompt Token Map.} The essential design of the proposed \emph{STA-VPT} for image prompting lies in the spatial alignment between the 2D prompt token map and the image token map, which is in contrast to the unordered sequential prompt structure in the sequential modeling paradigm. As shown in Figure~\ref{fig:framework}, the learned prompt token map has the same size as the image token map, or a scaled version of it. Besides, the prompt map shares the pre-learned positional embeddings from the pre-trained model with the image token map. Based on these designs, the established spatial alignment between the prompt and the visual tokens enables the learned prompt map to preserve the spatial structure and thereby capture the spatial relations within the input image. Moreover, it allows our \emph{STA-VPT} to perform fine-grained, individualized prompting for different image tokens, with each prompt token exclusively focusing on prompting its spatially aligned image token.

To minimize the parameter learning overhead, we learn a low-dimensional parameterized vector for each prompt token in Prompt pathway, and project these prompt tokens to a high-dimensional feature space that matches the feature dimension of the image tokens in Base pathway. Note that we perform linear projection for all prompt tokens using the same linear layer, incurring a negligible amount of learnable parameters, as illustrated in Figure~\ref{fig:framework}.


\smallskip\noindent\textbf{Fine-grained Spatial Prompting by Bilateral Interaction.} 
Our \emph{STA-VPT} performs fine-grained individual prompting by spatially aligned bilateral interaction (denoted as \emph{S-BI}) between two pathways. To be specific, Prompt pathway distills the relevant knowledge to the downstream task from Base pathway while Base pathway acquires the prompted knowledge by attending to Prompt pathway. 
As illustrated in Figure~\ref{fig:framework}, two  directions of bilateral interaction are performed asynchronously across the network depth to refine two pathways iteratively. Both directions of interactions are implemented using the designed spatially-aligned cross attention operation, where the interaction direction is reversed by exchanging the query with the key and value. 

\smallskip\noindent\textbf{Spatially-aligned Cross Attention.} Typical cross attention performs global interactions between a query token and all corresponding key and value tokens in each attention operation. In contrast, the proposed spatially-aligned cross attention restricts the query token to attend only to the value (key) tokens within its spatially aligned local area. Taking the cross attention from Base pathway to Prompt pathway as an example, a token located at $\langle x, y \rangle$ in the $l$-th layer of Base pathway serves as the query $\mathbf{q}^{\mathcal{B},l}_{\langle x, y \rangle}$ and attends to the tokens in the spatially corresponding locations in the $l$-th layer of Prompt pathway which is a local $c \times c$ squared window $\mathbf{R}^{\mathcal{P},l}$ centered at $\langle x, y \rangle$:
\begin{equation}
\begin{split}
    & \mathbf{q} = \mathbf{q}^{\mathcal{B},l}_{\langle x, y \rangle} \mathbf{W}^q, \ \mathbf{K} = \mathbf{R}^{\mathcal{P},l} \mathbf{W}^K, \ \mathbf{V} = \mathbf{R}^{\mathcal{P},l} \mathbf{W}^V, \\
    & \mathbf{o}^{\mathcal{B},l}_{\langle x, y \rangle} = \text{softmax}(\frac{\mathbf{q} \mathbf{K}^\top}{\sqrt{d_s}}) \mathbf{V} + \mathbf{q}^{\mathcal{B},l}_{\langle x, y \rangle}.
\end{split}
\label{eqn:CA}
\end{equation}
Herein, $\mathbf{R}^{\mathcal{P},l}$ denotes the set of tokens in the squared window of size $c \times c$ centered at $\langle x, y \rangle$ in the $l$-th layer of Prompt pathway while $d_s$ is a scaling factor. Residual connection is employed after the cross attention. Note that all the involved parameters in Equation~\ref{eqn:CA}, including $\mathbf{W}^q$, $\mathbf{W}^K$ and $\mathbf{W}^V$, are reused from the self-attention operation in the $l$-th layer of the pre-trained vision model, which eliminates the learning overhead. $c$ is a hyper-parameter defining the attention window size to balance between the local specificity and the global smoothness for prompting. $c=1$ corresponds to the most fine-grained prompting.

\smallskip\noindent\textbf{Prompt Adapter.} To facilitate the feature adaptation of prompted knowledge from the pre-trained backbone model to the downstream task, we additionally insert a lightweight prompt adapter after the cross attention operation. As shown in Figure~\ref{fig:framework}, the prompt adapter is implemented as a standard vanilla adapter~\cite{ref16}, consisting of two linear layers with a GeLU~\cite{hendrycks2016gaussian} activation layer and a normalization layer in between. Thus, the prompted feature at $\langle x, y \rangle$ is adapted by:
\begin{equation}
    \mathbf{o}^{\mathcal{B},l}_{\langle x, y \rangle}=\text{GeLU}(\text{LN}(\mathbf{o}^{\mathcal{B},l}_{\langle x, y \rangle})\cdot W_\text{down})\cdot W_\text{up},
\end{equation}
where the feature dimension is first shrunk from $d$ to $t$ ($t\ll d$) by the learnable matrix $W_{down}\in R^{d\times t}$ and then expanded back to $d$ by $W_{up}\in R^{t\times d}$. Inspired by the effective feature scaling technique~\cite{lian2022scaling} for feature adaptation, we perform dimension-wise scaling on the adapted prompt feature by learning a parameterized scaling vector $\mathbf{s} \in \mathbb{R}^{d}$, and fuse it with the original output features $\mathbf{f}^{\mathcal{B},l}_{\langle x, y \rangle}$ in the $l$-th layer of the backbone:
\begin{equation}
    \mathbf{f}^{\mathcal{B},l}_{\langle x, y \rangle} = \mathbf{s} \odot \mathbf{o}^{\mathcal{B},l}_{\langle x, y \rangle} + \mathbf{f}^{\mathcal{B},l}_{\langle x, y \rangle},
\end{equation}
where $\odot$ denotes the element-wise product between two vectors. Note that this dimension-wise feature scaling was not used in the conference version of this work~\cite{conf_version}.



\begin{figure*}[t]%
\centering
\includegraphics[width=0.98\linewidth]{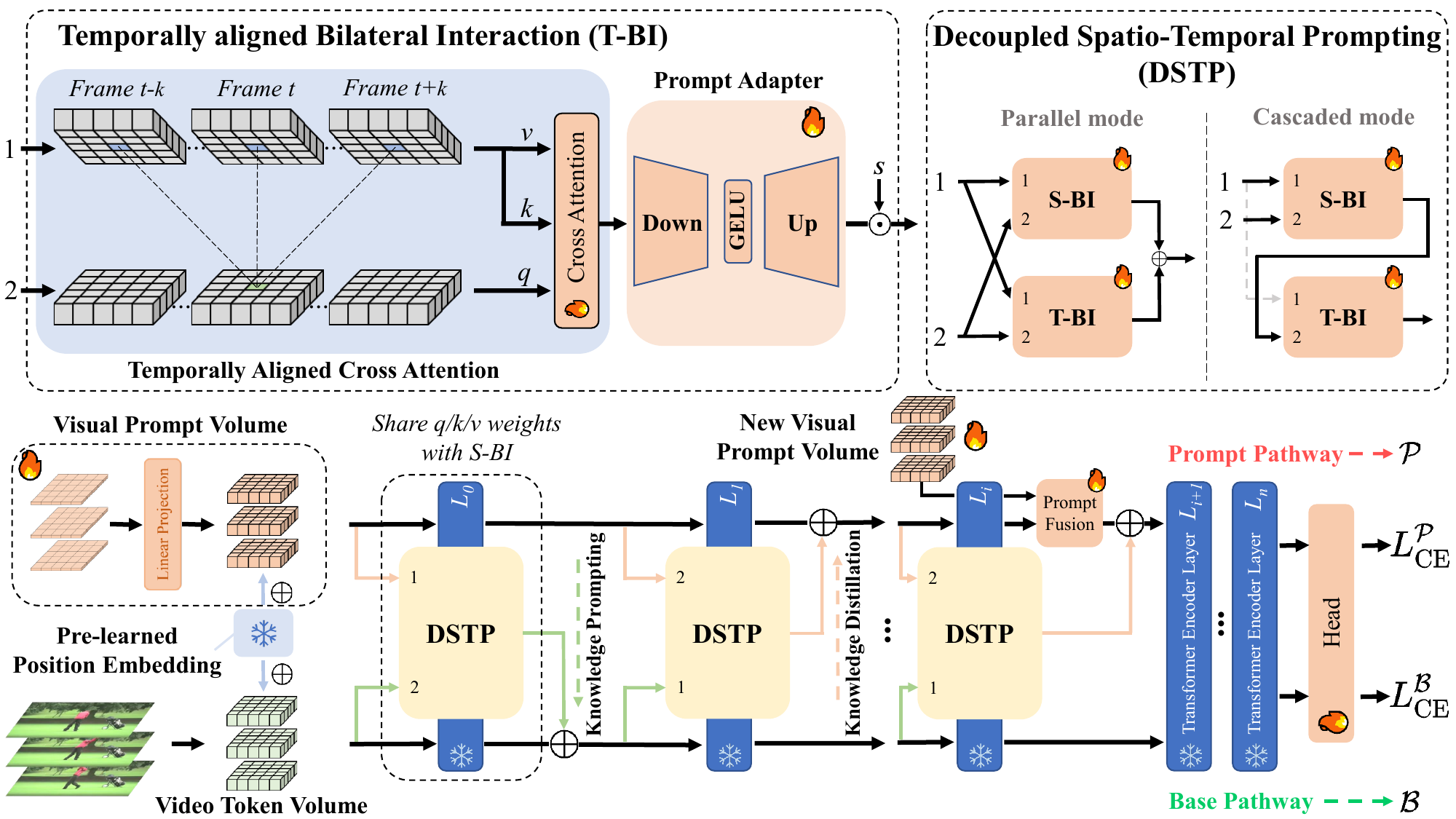}
\caption{Framework of the proposed \emph{STA-VPT} for video prompting with spatiotemporal alignment. \emph{STA-VPT} learns a 3D token volume which is spatiotemporally aligned with the video token volume. Bilateral interactions are applied in both spatial (S-BI presented in Figure~\ref{fig:framework}) and temporal domains (T-BI) to perform spatial and temporal prompting, respectively. The prompted knowledge in two domains are then integrated in either parallel or cadcaded modes for video prompting.}
\label{fig:framework_addtp}
\end{figure*}%

\smallskip\noindent\textbf{Inter-layer Prompt Fusion.} Similar to VPT, we can either learn the visual prompts only in the first layer termed as `shallow mode' or learn prompts at multiple ViT layers in `deep mode'. In general, more layers of visual prompts in the deep mode tend to outperform the shallow mode due to more learning capacity, albeit at the cost of increased parameter learning overhead. Besides, visual prompts at different layers can potentially distill and prompt knowledge across multiple levels of semantics corresponding to features at varying depths of the backbone network. In our deep mode, we learn prompts at three layers including the Layer 0, 4, and 8 of the backbone, with these positions set empirically. 


We also perform the dimension-wise feature scaling~\cite{lian2022scaling} on the newly learned prompt token map before fused into Prompt pathway to bridge the feature gap between them. As shown in Figure~\ref{fig:framework}, for a newly learned prompt token map $\mathbf{P}^{l} \in \mathbb{R}^{h^l \times w^l \times d}$ in the $l$-th layer of the backbone, we learn a parameterized scaling vector $\mathbf{s} \in \mathbb{R}^d$ and duplicate it spatially to have equal spatial size ($h^l \times w^l$) with $\mathbf{P}^{l}$, denoted as $\mathbf{S}\in \mathbb{R}^{h^l \times w^l \times d}$. Then we perform Hadmard product between $\mathbf{S}$ with $\mathbf{P}^{l}$ for feature scaling. The scaled results are added to the feature map $\mathbf{O}^{\mathcal{P},l}$ in the $l$-th layer of Prompt pathway for inter-layer prompt fusion:
\begin{equation}
    \mathbf{O}^{\mathcal{P},l} =\mathbf{S} \odot \mathbf{P}^{l} + \mathbf{O}^{\mathcal{P},l}.
\end{equation}

\subsubsection{Prompt Learning}
Following the experimental setting of VPT~\cite{ref28}, we apply our \emph{STA-VPT} to image prompting on two classical types of downstream vision tasks: image classification and semantic segmentation, corresponding to image-level and pixel-level prediction tasks, respectively.

\noindent\textbf{Prompt Learning for Image Classification.} 
The output features from Base pathway are ultimately used for image classification during inference, thus we primarily perform Cross-Entropy loss on Base pathway for supervision. Besides, we also apply Cross-Entropy loss to Prompt pathway as an auxiliary loss, which can potentially enhance the capability of the visual prompts to learn task-relevant knowledge for downstream tasks. Thus, our model is supervised by:
\begin{equation}
    L_\text{cls}=\lambda L_\text{CE}^{\mathcal{B}}(y_b,y^{*})+(1-\lambda) L_\text{CE}^{\mathcal{P}}(y_p,y^{*}),
\end{equation}
where $y_p$ and $y_b$ are predictions by Prompt pathway and Based pathway respectively while $y^{*}$ is the corresponding groundtruth. $\lambda$ is a hyper-parameter to balance two losses.

\smallskip\noindent\textbf{Prompt Learning for Semantic Segmentation.} Consistent with VPT, we utilize SETR~\cite{zheng2021rethinking}, a prominent segmentation model based on Transformer architecture, as the pre-trained backbone in this setting. Specifically, SETR consists of a feature encoder built upon ViT-L and a lightweight decoder used as a segmentation head. We apply our \emph{STA-VPT} to the encoder for visual prompt tuning. The parameters of encoder is frozen while only the prompting parameters and the decoder remain tunable. Similar to the supervised prompt learning on the classification task, we conduct supervision primarily on Base pathway while applying auxiliary supervision to Prompt pathway. To be specific, we use the same segmentation loss as SETR, i.e., pixel-level Cross-Entropy loss denoted as $L_\text{pixel-CE}$ on both pathways:
\begin{equation}
    L_\text{seg}=\alpha L_\text{pixel-CE}^{\mathcal{B}}(\mathbf{M}_b,\mathbf{M}^{*})+(1-\alpha) L_\text{pixel-CE}^{\mathcal{P}}(\mathbf{M}_p,\mathbf{M}^{*}),
\end{equation}
where $\mathbf{M}_b$ and $\mathbf{M}_p$ are predicted segmentation masks by Base pathway and Prompt pathway respectively. $\mathbf{M}^*$ is the segmentation groundtruth while $\alpha$ is tuned to balance two losses.

\subsection{Video Prompting with SpatioTemporal Alignment}
Effective feature learning for video data requires not only encoding individual image frames in the spatial domain but also capturing dynamic features across the temporal domain. To this end, our \emph{STA-VPT} performs video prompting in both the spatial and temporal domains. We utilize pre-trained CLIP-ViT-B~\cite{radford2021learning} as the video prompt tuning backbone for our \emph{STA-VPT}, which is consistent with the ViT backbone structure in image prompting. Accordingly, the video prompting version of our \emph{STA-VPT} adheres to the same modeling framework with image prompting, with an extension to the temporal domain to accommodate video modality, as shown in Figure~\ref{fig:framework_addtp}.  

\smallskip\noindent\textbf{SpatioTemporally Aligned Prompt Token Volume.} The proposed \emph{STA-VPT} learns a three-dimensional token volume in video prompting, which is aligned with the video token volume in both the spatial and temporal domains. This 3D-dimensional alignment enables the video prompts to learn the spatiotemporal relations in the input video. Meanwhile, \emph{STA-VPT} is able to perform fine-grained individualized prompting for video tokens in different spatiotemporal regions with spatiotemporal alignment to capture region-specific semantic patterns.

\smallskip\noindent\textbf{Decoupled SpatioTemporal Fine-grained Prompting.} Learning spatial and temporal features in a decoupled manner for video data has been shown to outperform joint learning of spatiotemporal features, as demonstrated by prominent methods such as SlowFast model~\cite{Slowfast}, (2+1)D spatiotemporal convolutions~\cite{2+1DConvolution}, and other two-stream designs~\cite{TSN, Two-stream} for video feature learning. Motivated by this insight, our \emph{STA-VPT} adopts such a decoupled strategy for video prompting by performing spatial prompting and temporal prompting separately, as depicted in Figure~\ref{fig:framework_addtp}. 

\smallskip\noindent\textbf{Temporally-aligned Cross Attention.} Similar to the spatial prompting implemented by bilateral interaction in the spatial domain (\emph{S-BI}) between Prompt pathway and Base pathway, the temporal prompting is achieved through bilateral interaction in the temporal domain (\emph{T-BI}) between the two pathways. Specifically, it is implemented with the designed temporally aligned cross attention in both directions as shown in Figure~\ref{fig:framework_addtp}, where a query token in a pathway attends only to the temporally corresponding value (key) tokens in the other pathway. This enable our \emph{STA-VPT} to conduct fine-grained individualized prompting in the temporal domain. Taking the cross attention from Base pathway to Prompt pathway as an example, a token in the $l$-th layer of Base pathway, located at $\langle x, y \rangle$ in the $t$-th frame, serves as the query $\mathbf{q}^{\mathcal{B},l}_{\langle x, y, t \rangle}$ and attends to the temporally corresponding tokens $\mathbf{R}^{\mathcal{P},l}$ in the $l$-th layer of Prompt pathway, which are within a local temporal window of size $c$ centered at $t$ in the input video:
\begin{equation}
\begin{split}
    & \mathbf{q} = \mathbf{q}^{\mathcal{B},l}_{\langle x, y, t \rangle} \mathbf{W}^q, \ \mathbf{K} = \mathbf{R}^{\mathcal{P},l} \mathbf{W}^K, \ \mathbf{V} = \mathbf{R}^{\mathcal{P},l} \mathbf{W}^V, \\
    & \mathbf{o}^{\mathcal{B},l}_{\langle x, y, t \rangle} = \text{softmax}(\frac{\mathbf{q} \mathbf{K}^\top}{\sqrt{d_s}}) \mathbf{V}.
\end{split}
\label{eqn:CA_TEMP}
\end{equation}
Herein, $d_s$ is a scaling factor and $c$ is a hyper-parameter that balances local specificity and global temporal smoothness. Setting $c=1$ yields the most fine-grained temporal prompting.

As illustrated in Figure~\ref{fig:framework_addtp}, we investigate two schemes for integrating prompting in the spatial and temporal domains: either performing prompting in parallel and fusing the results by element-wise addition, or adopting a cascaded approach with spatial prompting first. The performance of both integration schemes are evaluated in the experiments. 

\smallskip\noindent\textbf{Video Prompt Learning.} We adopt action recognition as the downstream task for video prompt tuning. Consistent with the image prompting, we apply supervision with Cross-Entropy loss to both Base pathway and Prompt pathway for prompt learning. 

\begin{table*}[!t]
\caption{Image classification performance of different PEFT methods based on ViT backbone on VTAB-1k benchmark. The best results are highlighted in bold, and the second-best are underlined.}
\label{tab:table_vtab}
\centering
\scriptsize
\tabcolsep=2pt
\renewcommand\arraystretch{1.0} 
\resizebox{1.0\linewidth}{!}{
\begin{tabular}{l|l|ccccccc@{\hspace{6.0pt}}c|cccc@{\hspace{6.0pt}}c|cccccccc@{\hspace{6.0pt}}c|cc} 
\toprule
\multicolumn{2}{c|}{Methods} & \multicolumn{8}{c|}{Natural} & \multicolumn{5}{c|}{Specialized} & \multicolumn{9}{c|}{Structured} & \multicolumn{2}{c}{}\\
\midrule
& & \rotatebox{90}{CIFAR-100} & \rotatebox{90}{Caltech101} & \rotatebox{90}{DTD} & \rotatebox{90}{Flowers102} & \rotatebox{90}{Pets} & \rotatebox{90}{SVHN} & \rotatebox{90}{Sun397} & \rotatebox{90}{\bf{Mean}} & \rotatebox{90}{Patch Camelyon} & \rotatebox{90}{EuroSAT} & \rotatebox{90}{Resisc45} & \rotatebox{90}{Retinopathy} & \rotatebox{90}{\bf{Mean}} & \rotatebox{90}{Clevr/count} & \rotatebox{90}{Clevr/distance} & \rotatebox{90}{DMLab} & \rotatebox{90}{KITTI/distance} & \rotatebox{90}{dSprites/location} & \rotatebox{90}{dSprites/orientation} & \rotatebox{90}{SmallNORB/azimuth} & \rotatebox{90}{SmallNORB/elevation} & \rotatebox{90}{\bf{Mean}} & \rotatebox{90}{\bf{Overall Mean}} & \rotatebox{90}{\bf{Tunable Param. (M)}}\\
\cmidrule{1-26}
Full Fine-tune & \multirow{5}{*}{Fine-tuning} &68.9 &87.7 &64.3 &97.2 &86.9 &87.4 &38.8 &75.88 &79.7 &95.7 &84.2 &73.9 &83.36 &56.3 &58.6 &41.7 &65.5 &57.5 &46.7 &25.7 &29.1 &47.64 &68.96 &85.8 \\
Head Fine-tune & &63.4 &85.0 &63.2 &97.0 &86.3 &36.6 &51.0 &68.93 &78.5 &87.5 &68.6 &74.0 &77.16 &34.3 &30.6 &33.2 &55.4 &12.5 &20.0 &9.6 &19.2 &26.84 &57.64 &0.04\\
BIAS & &72.8 &87.0 &59.2 &97.5 &85.3 &59.9 &51.4 &73.30 &78.7 &91.6 &72.9 &69.8 &78.25 &61.5 &55.6 &32.4 &55.9 &66.6 &40.0 &15.7 &25.1 &44.10 &65.22 & 0.11\\
GPS & &81.1 &\textbf{94.2} &\textbf{75.8} &\underline{99.4} &91.7 &\textbf{91.6} &52.4 &83.74 &\underline{87.9} &96.2 &86.5 &\underline{76.5} &\underline{86.78} &79.9 &62.6 &\textbf{55.0} &82.4 &84.0 &55.4 &29.7 &46.1 &61.89 &77.47 &$-$\\
TR-PTS & &81.2 &93.9 &\underline{75.1} &\textbf{99.5} &\textbf{91.9} &91.0 &54.5 &83.87 &\textbf{88.1} &95.7 &\textbf{87.8} &\textbf{76.6} &\textbf{87.05} &\underline{83.5} &63.2 &\underline{54.8} &\textbf{82.8} &\underline{87.7} &\textbf{56.9} &31.8 &46.1 &\underline{63.35} &\underline{78.09} &$-$\\
\midrule
Houlsby Adapter &\multirow{4}{*}{Adapter Tuning} &74.1 &86.1 &63.2 &97.7 &87.0 &34.6 &50.8 &70.50 &76.3 &88.0 &73.1 &70.5 &76.98 &45.7 &37.4 &31.2 &53.2 &30.3 &25.4 &13.8 &22.1 &32.39 &59.96 &0.27\\
AdaptFormer & &70.8 &91.2 &70.5 &99.1 &90.9 &86.6 &54.8 &80.56 &83.0 &95.8 &84.4 &76.3 &84.88 &81.9 &64.3 &49.3 &80.3 &76.3 &45.7 &31.7 &41.1 &58.83 &74.76 &0.19\\
SPT-Adapter & &72.9 &93.2 &72.5 &99.3 &91.4 &88.8 &55.8 &81.99 &86.2 &96.1 &85.5 &75.5 &85.83 &83.0 &\underline{68.0} &51.9 &81.2 &82.4 &51.9 &31.7 &41.2 &61.41 &76.41 &0.43\\
Adapter+ & &\underline{85.4} &93.8 &72.7 &99.1 &90.7 &88.2 &\textbf{58.1} &84.00 &87.5 &\underline{96.8} &\textbf{87.8} &73.9 &86.50 &83.4 &60.9 &53.8 &80.3 &87.2 &55.3 &\underline{37.9} &47.7 &63.31 &77.94 &0.27\\
\midrule
LoRA &\multirow{3}{*}{Reparameterization} &67.1 &91.4 &69.4 &98.8 &90.4 &85.3 &54.0 &79.49 &84.9 &95.3 &84.4 &73.6 &84.55 &82.9 &\textbf{69.2} &49.8 &78.5 &75.7 &47.1 &31.0 &44.0 &59.78 &74.61 &0.29\\
SSF & &69.0 &92.6 &\underline{75.1} &\underline{99.4} &\underline{91.8} &90.2 &52.9 &81.57 &87.4 &95.9 &87.4 &75.5 &86.55 &75.9 &62.3 &53.3 &80.6 &77.3 &54.9 &29.5 &37.9 &58.96 &75.69 &0.24\\
SPT-LoRA & &73.5 &93.3 &72.5 &99.3 &91.5 &87.9 &55.5 &81.93 &85.7 &96.2 &85.9 &75.9 &85.93 &\textbf{84.4} &67.6 &52.5 &82.0 &81.0 &51.1 &30.2 &41.3 &61.26 &76.37 & 0.63\\
\midrule
VPT-deep &\multirow{10}{*}{Prompt tuning} &78.8 &90.8 &65.8 &98.0 &88.3 &78.1 &49.6 &78.48 &81.8 &96.1 &83.4 &68.4 &82.43 &68.5 &60.0 &46.5 &72.8 &73.6 &47.9 &32.9 &37.8 &54.98 &71.96 &0.64\\
EXPRES & &78.0 &89.6 &68.8 &98.7 &88.9 &81.9 &51.9 &79.69 &84.8 &96.2 &80.9 &74.2 &84.03 &66.5 &60.4 &46.5 &77.6 &78.0 &49.5 &26.1 &35.3 &54.99 &72.90 &0.98\\
$\text{E}^{2}$VPT & &78.6 &89.4 &67.8 &98.2 &88.5 &85.3 &52.3 &80.01 &82.5 &\underline{96.8} &84.8 &73.6 &84.43 &71.7 &61.2 &47.9 &75.8 &80.8 &48.1 &31.7 &41.9 &57.39 &73.94 &0.31\\
SPT-VPT & &79.3 &92.6 &73.2 &\textbf{99.5} &91.0 &89.1 &51.2 &82.27 &85.4 &\underline{96.8} &84.9 &74.8 &85.48 &70.3 &64.8 &54.2 &75.2 &79.3 &49.5 &36.5 &41.5 &58.91 &75.55 &0.22\\
VAPT & &80.8 &91.9 &69.7 &98.8 &89.2 &86.7 &52.9 &81.43 &84.4 &96.5 &85.1 &74.5 &85.13 &74.8 &63.6 &50.0 &77.2 &86.1 &48.3 &33.8 &40.9 &59.34 &75.30 &0.29\\
DA-VPT & & $-$ & $-$ & $-$ & $-$ & $-$ & $-$ & $-$ & 80.25 & $-$ & $-$ & $-$ & $-$ & 85.12 & $-$ & $-$ & $-$ & $-$ & $-$ & $-$ & $-$ & $-$ & 58.71 & 74.69 & 0.21\\
DA-VPT+BIAS & &74.4 &92.7 &74.3 &\underline{99.4} &91.3 &\underline{91.5} &50.3 &81.99 &86.2 &96.2 &87.2 &76.3 &86.48 &81.3 &62.5 &52.8 &65.3 &84.9 &51.0 &33.1 &48.7 &59.95 &76.14 &0.24\\
CVPT & &81.5 &91.5 &74.0 &99.2 &91.4 &90.7 &54.5 &83.26 &85.8 &96.5 &\underline{87.6} &75.8 &86.43 &79.0 &67.2 &50.6 &\underline{82.7} &81.5 &53.0 &34.4 &45.3 &61.71 &77.13 &0.43\\
PRO-VPT & &84.5 &\underline{94.1} &73.2 &\underline{99.4} &\underline{91.8} &88.2 &57.2 &\underline{84.06} &87.7 &\underline{96.8} &86.6 &75.5 &86.65 &78.8 &61.0 &50.6 &81.3 &86.7 &\underline{56.4} &\textbf{38.1} &\underline{51.7} &63.08 &77.93 &0.61\\
\textbf{STA-VPT (Ours)} & & \textbf{85.7} & 93.7 & 73.7 & 99.1 & \textbf{91.9} & 88.7 & \underline{57.5} & \textbf{84.33} & 87.7 & \textbf{97.0} & 86.8 & 74.0 & 86.38 & 83.4 & 60.9 & 52.9 & 81.4 & \textbf{91.2} & 55.3 & 37.5 & \textbf{53.2} & \textbf{64.48} & \textbf{78.40} & 0.38\\
\bottomrule
\end{tabular}
}
\end{table*}

\section{Experiments}

We apply the proposed \emph{STA-VPT} to both image prompting and video prompting to evaluate its performance comprehensively. Specifically, we adopt image classification and semantic segmentation as the instantiated downstream tasks for image prompting, corresponding to image-level and pixel-level prediction tasks, respectively. For video prompting, we conduct experiments on action recognition, a classical vision task on video data. Additionally, we perform extensive ablation studies in the experiments of both image classification and action recognition to investigate the effectiveness of each essential design of \emph{STA-VPT}.

\subsection{Image Prompting for Image Classification}
\label{sec:exp-class}
\subsubsection{Experimental Setup}
\noindent\textbf{Datasets.} We conduct experiments of image classification on three standard benchmarks for parameter-efficient fine-tuning across diverse scenes: VTAB-1k~\cite{zhai2019large}, FGVC and HTA\cite{ref31}. 
VTAB-1k benchmark consists of $19$ datasets categorized to three groups of tasks: 1) `Natural' group that captures natural objects by standard cameras, 2) `Specialized' group taking photos utilizing specialized equipments like satellite or medical device, and 3) `Structured' group performing geometric comprehension such as counting and distance perception. Each dataset in VTAB-1k contains $1000$ training images, among which 800 images are for training and the left is for validation~\cite{zhai2019large}. 
FGVC benchmark contains $5$ image datasets, including CUB~\cite{ref33}, NABirds~\cite{ref34}, Oxford Flowers~\cite{ref35}, Stanford Dogs~\cite{ref36} and Stanford Cars~\cite{ref37}. We adopt the same data split as VPT for training and test. 
The head tuning adaptation benchmark (HTA)~\cite{ref31} comprises 10 datasets including CIFAR10~\cite{ref38}, CIFAR100 \cite{ref38}, DTD~\cite{ref39}, CUB-200~\cite{ref33}, NABirds~\cite{ref34}, Stanford-Dogs\cite{ref36}, Oxford-Flowers~\cite{ref35}, Food101~\cite{ref40}, GTSRB~\cite{ref41} and SVHN~\cite{ref42}. Following DAM-VP~\cite{ref31}, we use the default train-validate-test data split to have a fair comparison with other methods.

\noindent\textbf{Implementation Details.} Following VPT, we primarily use ViT-B/16 pre-trained on ImageNet-21K~\cite{deng2009imagenet} as the backbone. We also instantiate the backbone with a Swin Transformer~\cite{ref4} pre-trained on ImageNet-21K to assess robustness across different backbone architectures. During training, all the parameters of the backbone are frozen while only the prompt map and the classification head are tunable. 
AdamW~\cite{loshchilov2017decoupled} is used for optimization with the initial learning rate $1e^{-3}$, weight decay $1e^{-4}$ and the batch size 64 or 128. 
In this set of experiments, classification accuracy is used as evaluation metric.

\begin{table*}[!t]
\centering
\caption{Image classification performance of different PEFT methods based on ViT backbone on FVGC benchmark.}
\label{tab:fgvc}
\resizebox{1.0\linewidth}{!}{
\setlength{\tabcolsep}{1.3mm}{
\begin{tabular}{>{\small}l|>{\small}l|>{\small}c>{\small}c>{\small}c>{\small}c>{\small}c|>{\small}c>{\small}c} 
\toprule
\multicolumn{2}{c|}{Methods}  & CUB-200-2011 & NABirds & Oxford Flowers & Stanford Dogs & Stanford Cars & \textbf{Mean} & Tunable Param. (M)\\
\midrule
Full Fine-tune & \multirow{5}{*}{Fine-tuning} & 87.3 & 82.7 & 98.8 & 89.4 & 84.5 & 88.54 & 85.98\\
Head Fine-tune &  & 85.3 & 75.9 & 97.9 & 86.2 & 51.3 & 79.32 & 0.18\\
BIAS & & 88.4 & 84.2 & 98.8 & 91.2 & 79.4 & 88.40 & 0.28 \\
GPS & &89.9 &86.7 & \underline{99.7} &92.2 &90.4 &91.78 & $-$ \\
TR-PTS & &90.0 & 87.1 & 99.6 & \underline{92.4} & \underline{90.6} & \underline{91.94} & $-$ \\
\midrule
Houlsby Adapter & \multirow{4}{*}{Adapter tuning}& 87.1 & 84.3 & 98.5 & 89.8 & 68.6 & 85.67 & 0.41 \\
AdaptFormer &  & 84.7 & 75.2 & 97.9 & 84.7 & 83.1 & 85.12 & 0.37\\
SPT-Adapter &  & 89.1 & 83.3 & 99.2 & 91.1 & 86.2 & 89.78 & 0.40 \\
Adapter+ &  & 90.0	 & 83.2 & 99.6 & 91.6 & 89.1 & 90.70 & 0.34 \\
\midrule
LoRA & \multirow{3}{*}{Reparameterization}& 84.9 & 79.0 & 98.1 & 88.1 & 79.8 & 85.98 & 0.48\\
SPT-LoRA & & 88.6 & 83.4 & 99.5 & 91.4 & 87.3 & 90.04 & 0.52 \\
SSF & & 89.5 & 85.7 & 99.6 & 89.6 & 89.2 & 90.72 & 0.39 \\
\midrule
VPT-deep & \multirow{10}{*}{Prompt tuning}& 88.5 & 84.2 & 99.0 & 90.2 & 83.6 & 89.11 & 0.85\\
EXPRES & & 88.3 & $-$ & 99.0 & 90.0 & 80.5 & $-$ & $-$ \\
$\text{E}^{2}$VPT & & 89.1 & 84.6 & 99.1 & 90.5 & 82.8 & 89.22 & 0.56\\
SPT-VPT & & 90.6 & \underline{87.6} & \textbf{99.8} & 89.8 & 89.2 & 91.40 & 0.36 \\
VAPT & &89.7  &84.6  &99.1  &91.7  &82.8 & 89.58 & 0.67 \\
DA-VPT & & 90.2 & 87.4 & 99.4 & 89.4 & 89.7 & 91.22 & 0.30 \\
DA-VPT+BIAS & & \textbf{90.8} & \textbf{88.3} & \textbf{99.8} & 89.8 & \textbf{91.0} & \underline{91.94} & 0.32 \\
CVPT & & 89.9 & 86.5 & 99.3 & 91.7 & 85.2 & 90.52 & 0.77 \\
PRO-VPT & &90.6 &86.7 &\underline{99.7} &91.8 &89.6 &91.68 & 0.86 \\
\textbf{STA-VPT} (\textbf{Ours}) & & \underline{90.7} & 86.6 & \underline{99.7} & \textbf{92.5} & 90.5 & \textbf{92.00} & 0.87 \\
\bottomrule
\end{tabular}
}
}
\end{table*}

\begin{table*}[!t]
\centering
\small
\caption{Image classification performance of different PEFT methods based on ViT backbone on HTA benchmark.}
\label{tab:hta}
\resizebox{1.0\linewidth}{!}{
\setlength{\tabcolsep}{1.3mm}{
\begin{tabular}{>{\small}l|>{\small}l|>{\small}c>{\small}c>{\small}c>{\small}c>{\small}c>{\small}c>{\small}c>{\small}c>{\small}c>{\small}c|>{\small}c} 
\toprule
\multicolumn{2}{c|}{Methods} &DTD &CUB-200 &NABirds &Dogs &Flowers &Food-101 &CIFAR-100 &CIFAR-10 &GTSRB &SVHN & \textbf{Mean}\\
\cmidrule{1-13}
Full Fine-tune & \multirow{2}{*}{Fine-tuning} & 64.3 &87.3 &82.7 &89.4 &98.8 &84.9 &68.9 &97.4 &\textbf{97.1} &87.4 &85.82 \\
Head Fine-tune & &63.2 &85.3 &75.9 &86.2 &97.9 &84.4 &63.4 &96.3 &68.0 &36.6 &75.72\\
\midrule
Houlsby Adapter & \multirow{3}{*}{Adapter tuning} & 62.7 &87.1 &84.3 &89.8 &98.5 &86.0 &74.2 &97.7 &91.1 &36.3 &80.77\\
AdaptFormer & &74.4 &84.7 &75.2 &84.7 &97.9 &89.1 &91.4 &98.8 &\underline{97.0} &96.5 &88.97\\
Adapter+ & & 75.4 & 90.0 & 83.2 & 91.6 & 99.6 & \textbf{91.1} & 91.5 & 98.7 & 95.7 & \underline{97.3} & \underline{91.41}\\
\midrule
LoRA & \multirow{1}{*}{Reparameterization} & \textbf{77.0} &84.9 &79.0 &88.1 &98.1 & \underline{90.8} & \underline{91.8} & \underline{98.9} &96.5 &\underline{97.3} &90.24 \\
\midrule
VPT-deep & \multirow{5}{*}{Prompt tuning} & 65.8 &88.5 &84.2 &90.2 &99.0 &83.3 &78.8 &96.8 &90.7 &78.1 &85.54\\
DAM-VP & &73.1 &87.5 &82.1 &\underline{92.3} &99.2 &86.9 &88.1 &97.3 &90.6 &87.9 &88.50\\
$\text{E}^{2}$VPT & & 66.8 & 89.1 & 84.6 & 90.5 & 99.1 & 84.0 & 80.4 & 97.1 & 91.0 & 79.2 & 86.18 \\
SPT-VPT & & 66.5 & \underline{90.6} & \textbf{87.6} & 89.8 & \textbf{99.8} & 84.3 & 79.2 & 97.6 & 91.3 & 92.1 & 87.88\\
\textbf{STA-VPT (Ours)}& & \underline{76.4} & \textbf{90.7} & \underline{86.6} & \textbf{92.5} & \underline{99.7} & \textbf{91.1} & \textbf{91.9} & \textbf{99.0} & 96.8 & \textbf{97.5} & \textbf{92.22}\\
\bottomrule
\end{tabular}
}
}
\end{table*}

\noindent\textbf{Methods Involved in the Comparison.} We compare our \emph{STA-VPT} with state-of-the-art parameter-efficient fine-tuning methods that can be categorized into four types: 1) \textbf{standard fine-tuning} methods which either fine-tune all parameters of the backbone, termed as `Full Fine-Tune', or partially fine-tune the model including `Head Fine-Tune', `BIAS'~\cite{zaken2022bitfit}, `GPS'~\cite{zhang2024gradient} and `TR-PTS'~\cite{luo2025tr} ; 2) \textbf{adapter tuning} method, 
including vanilla Houlsby Adapter~\cite{ref16} and other more sophisticated methods such as `AdaptFormer'~\cite{ref21}, Apdater+~\cite{steitz2024adapters} and SPT-Adapter~\cite{he2023sensitivity}; 3) \textbf{visual prompt tuning} methods, including `VPT'~\cite{ref28} and its various refined variants, `E$^2$VPT'~\cite{han20232vpt}, `EXPRES'~\cite{ref57}, `DAM-VP'~\cite{ref31}, `VAPT'~\cite{le2026revisit}, `PRO-VPT'~\cite{shang2025pro} and SPT-VPT~\cite{wang2024revisiting}. Note that `VPT', `E$^2$VPT', `EXPRES', `VAPT', `PRO-VPT' and `SPT-VPT' follow the sequential prompting paradigm, whereas DAM-VP learns cluster-wise visual prompts to tackle the large dataset diversity of downstream tasks; 4) \textbf{reparameterization-based} methods, including LoRA~\cite{hu2022lora} which tunes partial parameters of the pre-trained model efficiently in a low-rank adaptation manner and SSF~\cite{lian2022scaling} that conducts parameterized dimension-wise feature affine transformation. 

\subsubsection{Comparison with Sequential Prompting Paradigm}
We first conduct a detailed comparison between our method and those following the sequential visual prompting paradigm. We present the experimental comparison based on ViT backbone between our \emph{STA-VPT} and other methods on VTAB-1k, FGVC and HTA datasets in Table~\ref{tab:table_vtab},~\ref{tab:fgvc} and~\ref{tab:hta}, respectively. While various enhanced variants of VPT exhibit varying degrees of improvements over original VPT, our \emph{STA-VPT} consistently outperforms all these visual prompting methods across all experiments, with particularly large margins on the VTAB-1k and HTA datasets. These results demonstrate the advantages of our method over the sequential prompting paradigm. Besides, Table~\ref{tab:vtab-swin} shows the comparison conducted using the Swin Transformer as the backbone. The large performance gains of our method over other visual prompting approaches further demonstrate the robustness of our method across different backbones. 

\begin{table*}[!t]
\centering
\caption{Image classification performance of different PEFT methods based on Swin Transformer backbone on VTAB-1k benchmark.\label{tab:vtab-swin}}
\renewcommand\arraystretch{1.0} 
\resizebox{0.7\linewidth}{!}{
\begin{tabular}{>{\small}l|>{\small}l|>{\small}c|>{\small}c>{\small}c>{\small}c>{\small}c} 
\toprule
\multicolumn{2}{c|}{\multirow{2}{*}{Methods}} & \multirow{2}{*}{Tuned/Total} & \multicolumn{4}{c}{VTAB-1k}\\
\cmidrule{4-7}
\multicolumn{2}{c|}{} & & Natural & Specialized & Structured & Mean \\
\cmidrule{1-7}
Full Fine-tune & \multirow{2}{*}{Fine-tuning} & 100\% & 79.10 & 86.21 & 59.65 & 74.99\\
Head Fine-tune & & 0.04\% & 73.52 & 80.77 & 33.52 & 62.60\\
\midrule
LoRA & \multirow{2}{*}{Reparameterization} & 1.18\% & 81.70 & 87.20 & 60.10 & 76.33\\
SPT-LoRA & & 0.49\% & 83.10 & \underline{87.40} & 60.40 & 76.97\\
\midrule
Houlsby Adapter & \multirow{3}{*}{Adapter tuning} & 1.18\% & 81.70 & 87.30 & 61.20 & 76.73\\
SPT-Adapter & & 0.33\% & 83.00 & 87.30 & 62.10 & 77.47\\
Adapter+ & & 0.40\% & \underline{83.68} & 87.32 & \underline{61.74} & \underline{77.58}\\
\midrule
VPT-deep & \multirow{3}{*}{Prompt tuning} & 0.23\% & 76.78 & 83.33 & 51.85 & 70.65\\
$\text{E}^{2}$VPT & & 0.21\% & 83.31 & 84.95 & 57.35 & 75.20\\
\textbf{STA-VPT (Ours)} & & 0.39\% & \textbf{83.95} & \textbf{87.46} & \textbf{62.69} & \textbf{78.03}\\
\bottomrule
\end{tabular}
}
\end{table*}

\subsubsection{Comparison with State-of-the-art Methods} We conduct quantitative comparison between our \emph{STA-VPT} and other \emph{PEFT} methods based on both `ViT' and `Swin Transformer' backbones, followed by more qualitative evaluation.

\noindent\textbf{Quantitative Evaluation with `ViT' Backbone.}
The results on VTAB-1k in Table~\ref{tab:table_vtab} show that our \emph{STA-VPT} achieves the best \emph{overall mean} performance on three groups of datasets. Particularly, `Structured' group is much more challenging than the other two groups since comprehending the structure information like counting or distance perception is generally more difficult than the classification of object categories. Our model surpasses other methods substantially on `Structured' group. 
Table~\ref{tab:fgvc} and Table~\ref{tab:hta} report the comparative results between all four categories of methods on FVGC and HTA benchmarks, respectively. Our \emph{STA-VPT} achieves the best \emph{mean} performance on both, further validating its effectiveness and robustness across diverse scenarios. Another noteworthy observation is that \emph{Adatper+} and \emph{LoRA} achieve competitive performance on HTA, surpassing all sequential prompt tuning methods. Nevertheless, our \emph{STA-VPT} performs even better, highlighting the advantage of the spatiotemporally aligned prompting of \emph{STA-VPT} over the sequential prompting paradigm. 

\noindent\textbf{Quantitative Evaluation with `Swin Transformer' Backbone.} 
Table~\ref{tab:vtab-swin} shows the comparison on VTAB-1k benchmark between our model and other methods when instantiating the backbone with Swin Transformer. The results show that our \emph{STA-VPT} outperforms other methods on all three groups of datasets, leading to the best \emph{mean} performance, which demonstrates the robustness of our \emph{STA-VPT} across backbones. 

\begin{figure*}[!t]%
\centering
\includegraphics[width=0.85\linewidth]{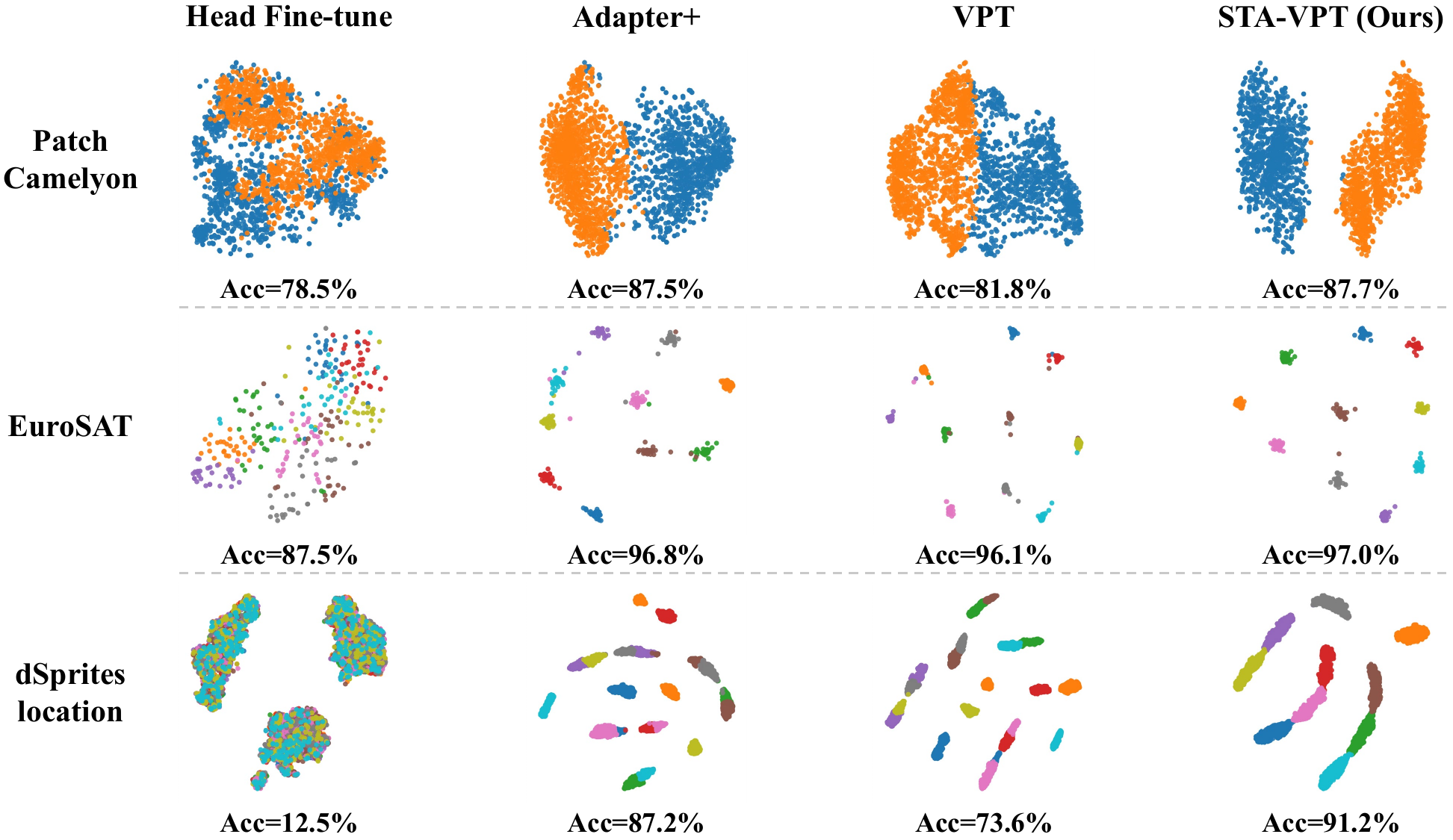}
\caption{t-SNE maps of learned features in the last layer of the backbone by four different methods on three datasets randomly selected from three groups of VTAB-1k benchmark. Our \emph{STA-VPT} exhibits more precise clustering w.r.t. different classes than other methods.}
\label{fig:tsne}
\end{figure*}

\begin{figure}[!t]%
\centering
\includegraphics[width=0.98\linewidth]{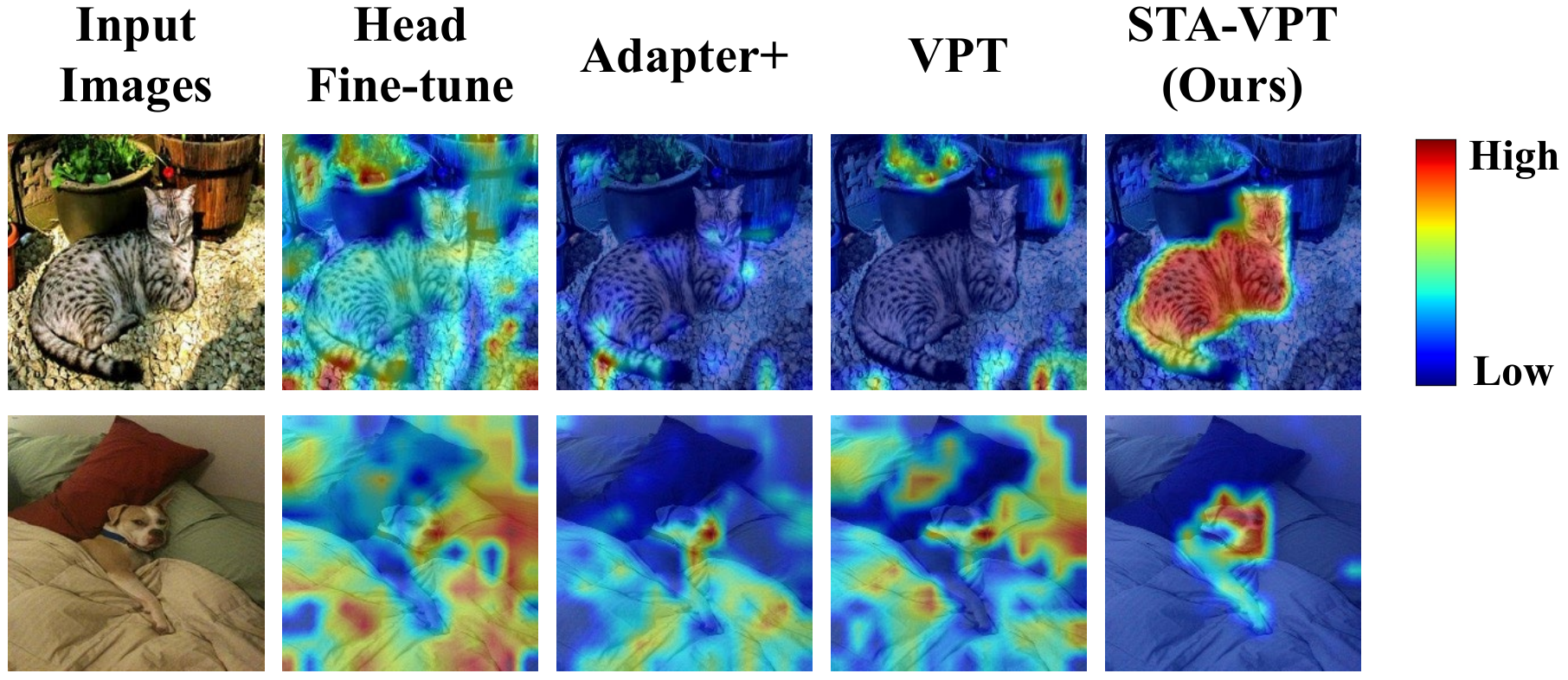}
\caption{GradCAM maps of features in the last layer of the backbone by different methods on two randomly selected samples. Our model is more focused on the labeled target objects i.e., `cat' and `dog' in these two samples, respectively.}
\label{fig:gradcam}
\end{figure}

\noindent\textbf{Qualitative Evaluation.} To evaluate the effect of prompt tuning on the feature learning for classification, we further conduct two sets of qualitative evaluation. First, we visualize the t-SNE~\cite{ref43} map of the extracted features in the last layer of the backbone (ViT) for different methods 
on randomly selected sub-datasets from VTAB-1k benchmark. Figure~\ref{fig:tsne} clearly show that our \emph{STA-VPT} has more precise clustering w.r.t. different classes than other methods, which reveals the higher quality of learned features by our model. Subsequently, we visualize the GradCAM~\cite{ref44} map of features in the last layer of the backbone for these methods, which indicates the attention regions of learned features by each method. The results in Figure~\ref{fig:gradcam} show that our \emph{STA-VPT} focuses more tightly on the target object than other methods. 

\subsubsection{Ablation Study}
To investigate the effectiveness of each core design and the tuning of key hyper-parameters, In this section we conduct a comprehensive ablation study on our \emph{STA-VPT} for image prompting, reporting results in Table~\ref{table:ablation}.

\noindent\textbf{Effect of Spatially Aligned Prompting.} To validate the effect of the spatially aligned prompting, we perform global cross attention instead of the spatially aligned cross attention during bilateral interaction, resulting in the uniform prompt variant which is essentially similar to VPT. The large performance gap between it (`w/o Spatial alignment') and the intact model across both benchmarks in Table~\ref{table:ablation} demonstrates the pivotal role of the spatially aligned prompting for STA-VPT.

\begin{table}[!t]
\centering
\caption{Ablation study of various core technical components of \emph{STA-VPT} on FVGC and VTAB-1k (`w/o' denotes ablation).}
\renewcommand\arraystretch{1.0} 
\resizebox{1.0\linewidth}{!}{
\begin{tabular}{>{\small}l|>{\small}c|>{\small}c>{\small}c>{\small}c>{\small}c} 
\toprule
\multirow{2}{*}{Ablated Variants} & \multirow{2}{*}{FVGC} & \multicolumn{4}{c}{VTAB-1k}\\
 & & Natural & Specialized & Structured & Mean\\
\midrule
w/o Spatial alignment & 90.58 & 83.19 & 83.06 & 60.51 & 75.59\\
w/o Prompt adapter   & 90.41 & 83.79 & 84.07 & 59.26 & 75.71\\
w/o Deep prompting   & 90.56 & 84.27 & 85.60 & 63.15 & 77.67\\
w/o Supervision on Prompt & 90.64 & 84.30 & 85.75 & 62.88 & 77.64\\
\midrule
Intact \emph{STA-VPT} & \textbf{92.00} & \textbf{84.33} & \textbf{86.38} & \textbf{64.48} & \textbf{78.40}\\
\bottomrule
\end{tabular}
\label{table:ablation}
}
\end{table}

\begin{figure}[!t]%
\centering
\includegraphics[width=0.96\linewidth]{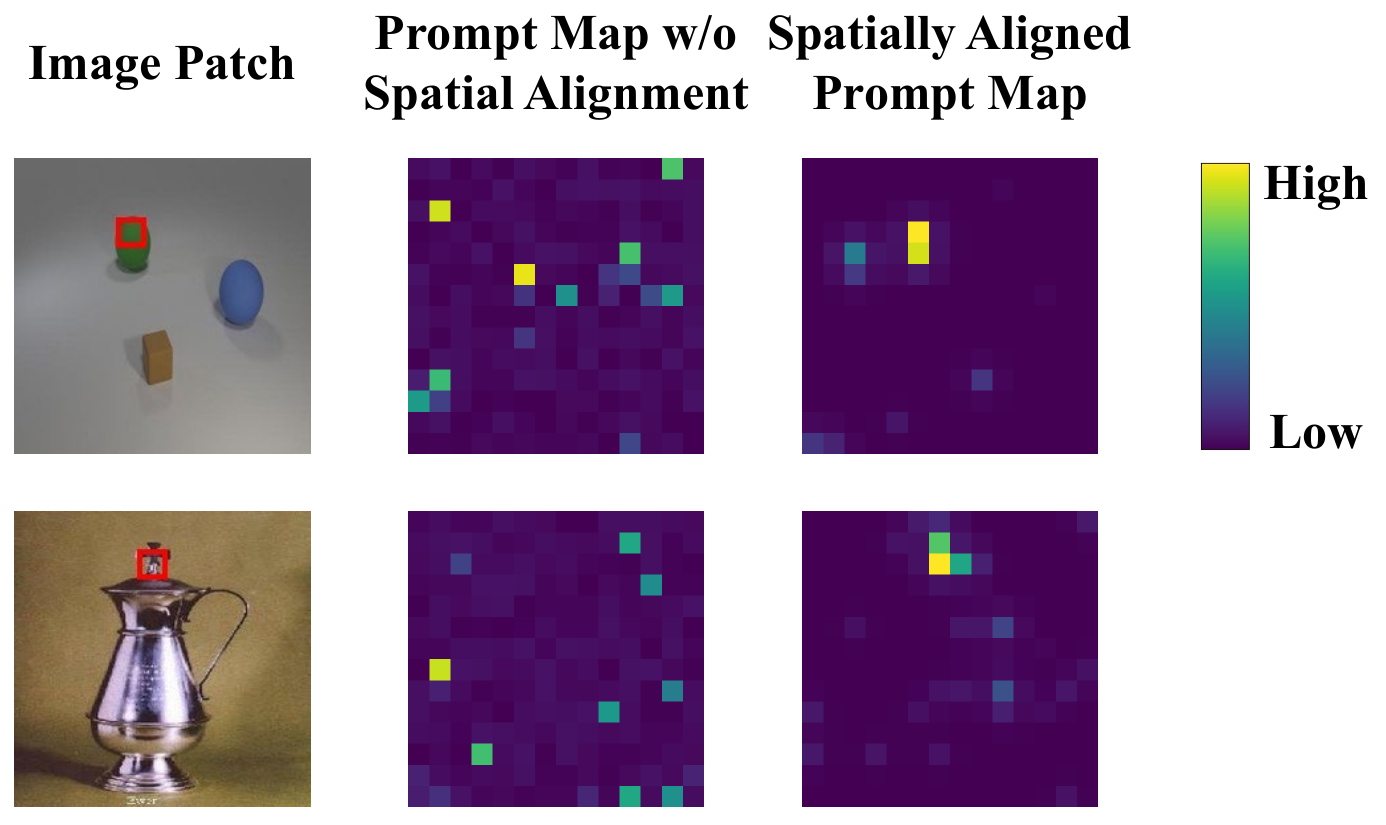}
\caption{Cosine similarity maps between a visually salient token in the image token map (red bounding box in the image patches) and all learned prompt tokens. Our model can indeed preserve such spatial alignment due to the proposed spatially aligned prompting paradigm.} 
\label{fig:similarity}
\end{figure}

To obtain more insight into the effect of the spatial alignment, we further visualize the similarity relations between the learned prompt map and the image token map. We randomly select a visually salient token on the object from the image feature map and calculate the Cosine similarities between the salient token and all tokens in the prompt map. We compare the result of our model with that of the uniform prompt variant(i.e. w/o spatial alignment). Figure~\ref{fig:similarity} indicate that our model can indeed preserve the spatial alignment between the learned prompt map and the image feature map, which is potentially beneficial to our model to capture spatial relations in the visual prompts.

\noindent\textbf{Effect of Prompt Adapter.} Table \ref{table:ablation} shows that the proposed prompt adapter is crucial to \emph{STA-VPT}’s performance on FGVC and all three VTAB-1k groups. This result confirms that the adapter effectively facilitates feature adaptation between the prompt and base pathways, thereby bridging the domain gap between the pre-trained model and the task-specific prompted knowledge.

\noindent\textbf{Deep vs Shallow Prompting Mode.} We also compare the performance of `deep' and `shallow' prompting modes in Table~\ref{table:ablation}, which reveals that the deep mode outperforms the shallow mode due to more thorough multi-layer prompting as well as larger learning capacity, consistent with \emph{VPT}. 

\noindent\textbf{Effect of Supervision on Prompt Pathway.} We supervise the prompt pathway to align the prompted knowledge with the downstream task objective. As shown in Table \ref{table:ablation}, this design is effective in most cases, with pronounced gains on the more challenging Structured group of VTAB-1k.



\begin{table*}[!t]
\centering
\caption{Semantic segmentation performance of different PEFT methods based on ViT backbone on ADE20K dataset. The best results are highlighted in bold, and the second-best are underlined.}
\renewcommand\arraystretch{1.0} 
\resizebox{0.65\linewidth}{!}{
\begin{tabular}{>{\small}l|>{\small}l|>{\small}c|>{\small}c|>{\small}c} 
\toprule
\multicolumn{2}{c|}{Methods} & mIoU-SS & mIoU-MS & Tunable Param. (M)\\
\cmidrule{1-5}
Full Fine-tune & \multirow{3}{*}{Fine-tuning} & \bf{48.31} & \bf{50.07} & 318.31\\
Head Fine-tune & & 39.87 & 42.17 & 14.16\\
BIAS & & 43.91 & 46.14 & 14.43\\
\midrule
LoRA & \multirow{2}{*}{Reparameterization} & 42.60 & 44.91 & 17.31\\
SPT-LoRA & & 45.40 & 47.50 & 14.60\\
\midrule
Houlsby Adapter & \multirow{3}{*}{Adapter} & 45.32 & 47.45 & 17.33\\
AdaptFormer & & 45.08 & 47.10 & 17.38\\
SPT-Adapter & & 45.20 & 47.20 & 14.60\\
\midrule
VPT-deep & \multirow{6}{*}{Prompt tuning} & 42.51 & 44.61 & 16.62\\
SPT-VPT & & 39.95 & 42.48 & 16.62\\
DA-VPT & & 45.10 & 47.07 & 13.60 \\
DA-VPT+BIAS & & 46.47 & 47.21 & 13.70 \\
CVPT & & 45.66 & 47.92 & 18.00\\
\textbf{STA-VPT (Ours)} & & \underline{46.77} & \underline{49.08} & 18.15\\
\bottomrule
\end{tabular}
}
\label{table_segm}
\end{table*}

\begin{table}[!t]
\caption{Effect of (i) the prompt-map size and (ii) spatially-aligned cross attention window size between two pathways on STA-VPT performance on VTAB-1k.}
\vspace{-4pt}
\label{table:map_size}
\centering
\tabcolsep=2pt
\renewcommand\arraystretch{1.0} 
\resizebox{0.95\linewidth}{!}{
\begin{tabular}{c|c|cccc} 
\toprule
\multirow{2}{*}{Map size} & \multirow{2}{*}{Window size}  & \multicolumn{3}{c}{VTAB-1k}\\
 & & Natural & Specialized & Structured & Mean \\
\midrule
\multirow{3}{*}{$4\times 4$} & $1$   &  82.76 & 85.75 & 60.90  & 76.47\\
 & $3$   & 83.66 & 85.21 & 63.31 & 77.39\\
 & $5$   & 83.14 & 85.92 & 63.00 & 77.35\\
 \midrule
\multirow{3}{*}{$7\times 7$} & $1$  & 83.55 & 85.29 & 62.20 & 77.01\\
 & $3$  & 84.06 & 85.70 & 63.46 & 77.74 \\
 & $5$  & 83.26 & 85.53 & 63.37 & 77.39 \\
 \midrule
\multirow{3}{*}{$14\times 14$} & $1$  & 84.10 & 85.74 & 63.26 & 77.70 \\
 & $3$  & \textbf{84.33} &\textbf{86.38} & \textbf{64.48} & \textbf{78.40}\\
 & $5$  & 84.32 & 85.50 & 63.59 & 77.80\\
\bottomrule
\end{tabular}
\vspace{-8pt}
}
\end{table}

\noindent\textbf{Tuning of prompt map size.} The prompt token map in our \emph{STA-VPT} is designed to be scalable while preserving spatial alignment with the image feature map. This introduces an efficiency-effectiveness trade-off: a smaller prompt map has fewer learnable parameters and is more efficient whereas its reduced capacity may degrade performance. Table~\ref{table:map_size} shows the performance of our model on VTAB-1k benchmark for different size of prompt token maps, validating the above analysis. While our model achieves competitive performance at $4\times 4$, the optimal setting is $14\times 14$, equal to the image feature-map size.

\noindent\textbf{Tuning of the spatially-aligned cross attention window size.} As defined in Equation~\ref{eqn:CA}, during spatially aligned cross-attention in \emph{STA-VPT}, each query token in one map (e.g., the image feature map) attends to a window in the other map (e.g., the prompt map) centered at its spatially aligned location. The window size $\mathbf{c}$ is a hyper-parameter to balance between the local specificity and the global smoothness in prompting. Tuning results in Table \ref{table:map_size} indicate that $\mathbf{c}=3$ yields the best performance.


\begin{figure*}[t]%
\centering
\includegraphics[width=1.0\linewidth]{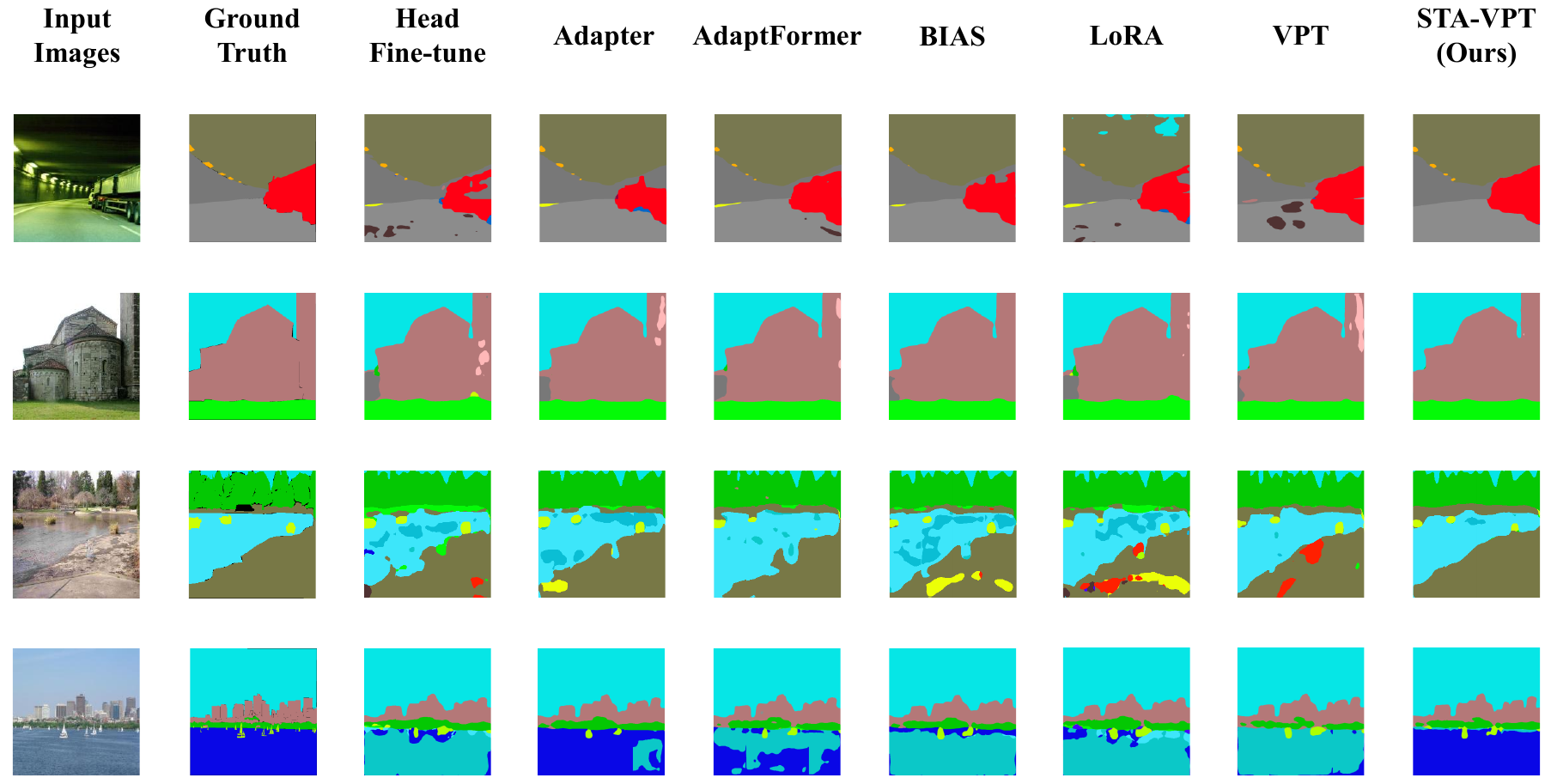}
\caption{Predicted segmentation masks by different PEFT methods on four challenging samples randomly selected from ADE20K validation set.}
\label{fig:vis_seg}
\end{figure*}%

\begin{figure*}[t]
    \centering
    \subfloat[\footnotesize Golf in HMDB51.]{
        \includegraphics[width=0.98\linewidth]{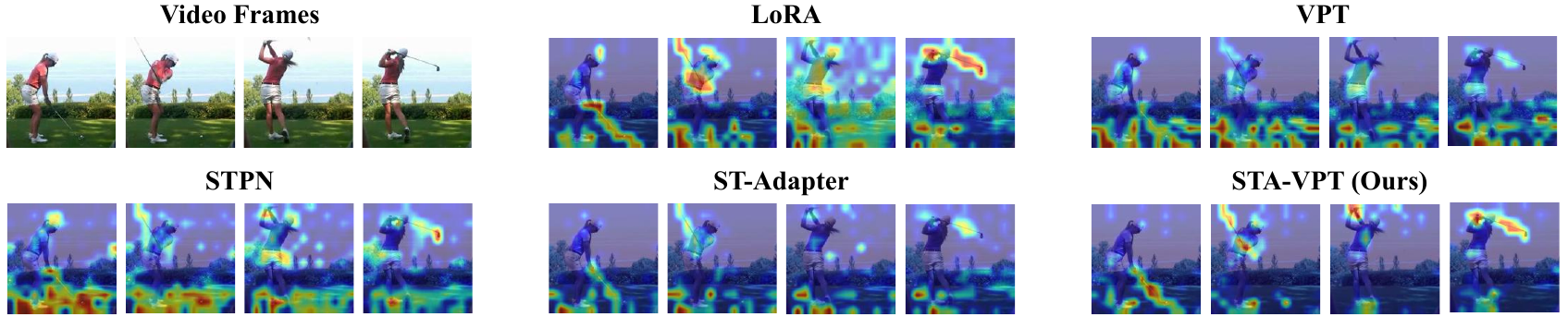}
        \label{fig:subfig-a}
    }
    
    \subfloat[\footnotesize Pretending to put something underneath something in SSv2.]{
        \includegraphics[width=0.98\linewidth]{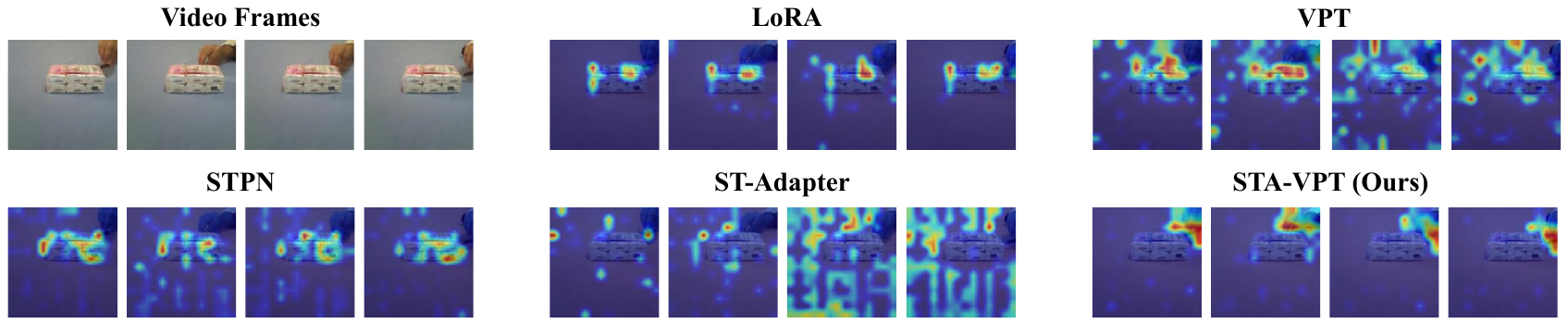}
        \label{fig:subfig-b}
    }
    
    \subfloat[\footnotesize HorseRiding in UCF101.]{
        \includegraphics[width=0.98\linewidth]{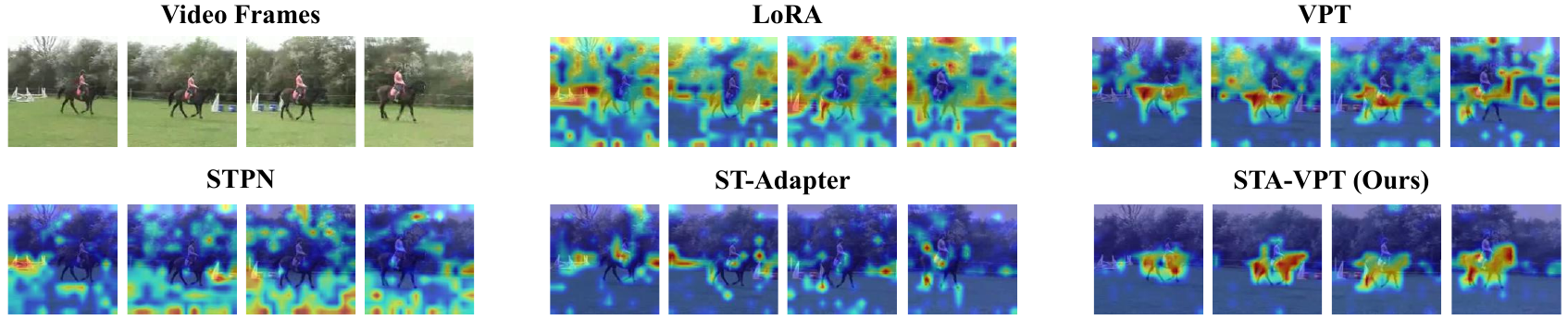}
        \label{fig:subfig-c}
    }
    
    \caption{GradCAM feature maps in the last layer of the backbone
by different methods on three randomly selected video samples from different datasets. Our \emph{STA-VPT} exhibits sharper and more accurate focus on action-relevant objects than other methods.}
    \label{fig:vis_video}
\end{figure*}

\begin{figure*}[t]%
\centering
\includegraphics[width=0.95\linewidth]{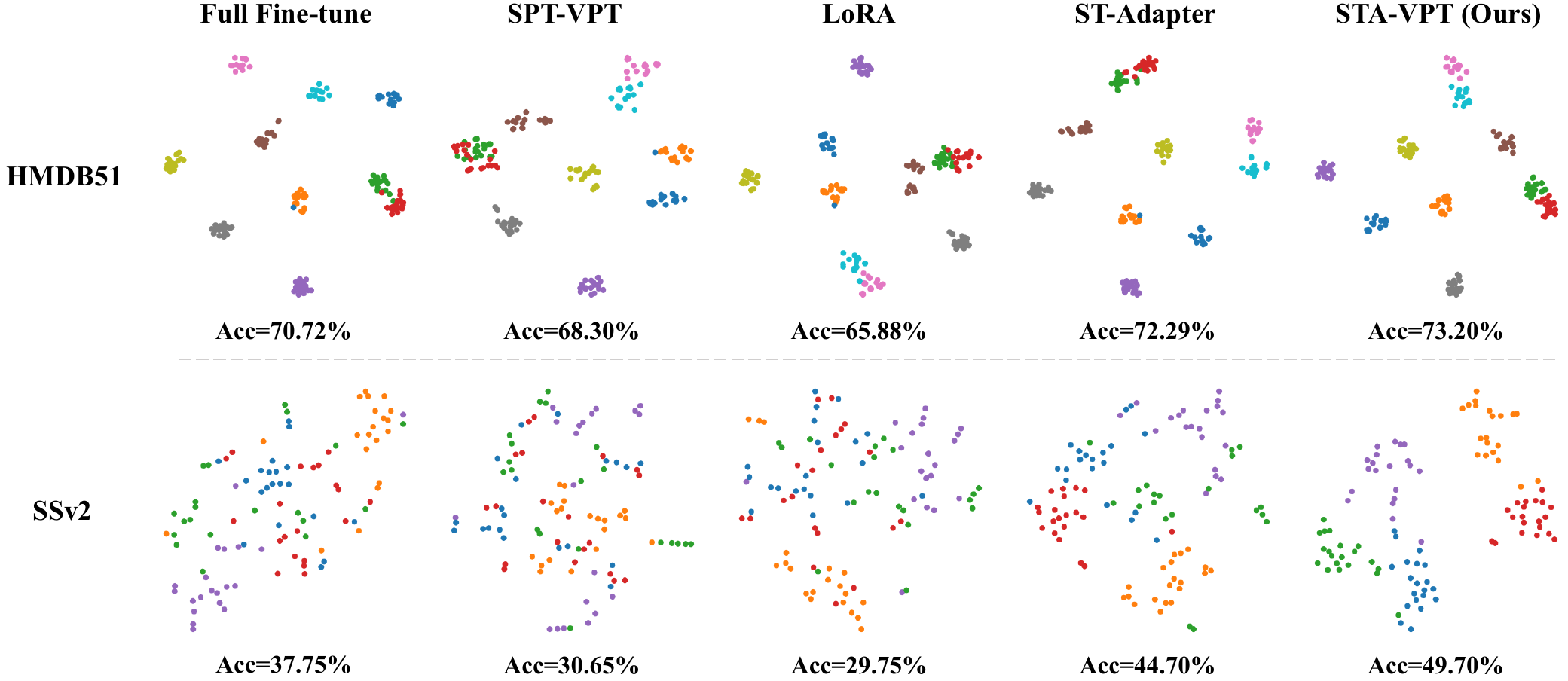}
\caption{T-SNE maps of last-layer backbone features for different PEFT methods on two randomly sampled subsets from HMDB51 and SSv2, respectively.}
\label{fig:video_tsne}
\end{figure*}%

\begin{table*}[!t]
\vspace{-6pt}
\centering
\caption{Action recognition performance of different PEFT methods based on ViT backbone across four datasets. We investigate two spatial–temporal prompting integration schemes—parallel and cascaded—denoted `STA-VPT-Parallel' and `STA-VPT-Cascaded', respectively.}
\label{tab:table_video}
\renewcommand\arraystretch{1.0} 
\resizebox{0.85\linewidth}{!}{
\begin{tabular}{>{\small}l|>{\small}l|>{\small}c|>{\small}c|>{\small}c|>{\small}c|>{\small}c|>{\small}c} 
\toprule
\multicolumn{2}{c|}{Methods} & HMDB51 & SSv2 & UCF101 &  K100 & Mean & Tunable Param. (M)\\
\cmidrule{1-8}
Full Fine-tune & \multirow{2}{*}{Fine-tuning} & 70.72 & 37.75 & 94.50 & 83.05 & 71.51 & 85.87\\
Head Fine-tune & & 55.49 & 20.80 & 86.41 & 77.15 & 59.96 & 0.07\\
\midrule
LoRA & \multirow{1}{*}{Reparameterization} & 65.88 & 29.75 & 91.81 & 81.90 & 67.34 & 2.43\\
\midrule
Houlsby Adapter & \multirow{4}{*}{Adapter} & 68.04 & 35.90 & 93.44 & 81.95 & 69.83 & 2.44\\
AdaptFormer & & 64.12 & 37.00 & 93.34 & 81.60 & 69.02 & 2.46\\
ST-Adapter	& & 72.29 & 44.70 & 94.90 & 84.05 & 73.99 & 3.04\\
AIM	& & 68.37 & 45.70 & 94.82 & 84.25 & 73.29 & 10.73\\
\midrule
VPT-deep & \multirow{5}{*}{Prompt tuning} & 70.33 & 34.70 & 94.26 & 84.50 & 70.95 & 4.68\\
SPT-VPT & & 68.30 & 30.65 & 93.15 & 84.35 & 69.11 & 4.68\\
STPN & & 71.70 & 33.70 & 92.97 & \underline{85.45} & 70.96 & 3.41\\
\textbf{STA-VPT-Parallel (Ours)} & & \bf{73.20} & \bf{49.70} & \bf{95.48} & \bf{85.70} & \textbf{76.02} & 5.35\\
\textbf{STA-VPT-Cascaded (Ours)} & & \underline{72.75} & \underline{46.30} & \underline{95.03} & \textbf{85.70} & \underline{74.95} & 5.35\\
\bottomrule
\end{tabular}
}
\end{table*}

\subsection{Image Prompting for Semantic Segmentation}

\subsubsection{Datasets and Implementation Details.} 
We conduct experiments on ADE20K~\cite{zhou2019semantic}, a challenging scene parsing benchmark with 150 fine-grained categories. The dataset includes 20,210 training images and 2,000 validation images.

Following \emph{VPT}, we employ SETR~\cite{zheng2021rethinking}, a ViT-L/16-based segmentation framework pre-trained on ImageNet-21K, as the backbone for our \emph{STA-VPT} in this set of experiments. We opt for progressive upsampling (PUP) decoding strategy and single/multi-scale inference for SETR, which is also consistent with \emph{VPT} configuration.  
`mIoU-SS' and `mIoU-MS' are used as the evaluation metrics for semantic segmentation in this set of experiments.


\subsubsection{Comparison with Sequential Prompting Paradigm}
We first compare our \emph{STA-VPT} with prompting tuning methods that follow the sequential modeling paradigm, to assess performance across the two visual prompting paradigms. The results falling in the `Prompt tuning' method category in Table~\ref{table_segm} indicate that our model substantially outperforms these methods. Specifically, our \emph{STA-VPT} surpasses \emph{VPT-deep} by 4.26 and 4.47 mIoU on single-scale (mIoU-SS) and multi-scale (mIoU-MS) settings, respectively. While \emph{DA-VPT} improves upon VPT by introducing auxiliary category constraints—and integrating \emph{BIAS} further enhances its performance—\emph{STA-VPT} still exceeds these methods by a large margin, demonstrating the superiority of spatiotemporally aligned visual prompt tuning over the sequential paradigm. 

\subsubsection{Comparison with State-of-the-art Methods}
\smallskip\noindent\textbf{Quantitative Evaluation.} Table \ref{table_segm} reports comprehensive results across all four categories of PEFT methods. \emph{Full fine-tuning} attains the best performance—presumably due to its greater adaptation capacity for dense prediction tasks like semantic segmentation—at the cost of substantially higher tuning overhead. While adapter-tuning methods perform on par with sequential prompt-tuning methods as well as \emph{SPT-LoRA}, our \emph{STA-VPT} markedly outperforms these PEFT baselines, validating its effectiveness.

\smallskip\noindent\textbf{Qualitative Evaluation.} We further provide a qualitative comparison by visualizing predicted segmentation masks from each PEFT method on four challenging samples randomly selected from the ADE20K validation set (Figure \ref{fig:vis_seg}). The results illustrate that our methods is able to produce more precise segmentation masks than other methods, especially in scenes with small objects and cluttered background.


\subsection{Video Prompting for Action Recognition}

\subsubsection{Experimental Setup}
\smallskip\noindent\textbf{Datasets.} To evaluate the video prompting performance of our \emph{STA-VPT}, we apply \emph{STA-VPT} with spatiotemporal alignment to action recognition across four standard benchmarks including HMDB51~\cite{HMDB51}, UCF101~\cite{UCF101}, SSv2 (small-scale)~\cite{SSv2} and Kinetics~\cite{Kinetics}. HMDB51 dataset contains 6,766 videos from 51 classes and UCF101 comprises 13320 videos from 101 classes. To simulate low-data learning scenarios, we use only a small set of training samples for prompt tuning on SSv2 and Kinetics, matching the scale used for HMDB51 and UCF101. Specifically, we randomly choose 100 classes and 80/20 videos per class from their original datasets for training/testing, For each video, we uniformly sample 8 frames (i.e., $T=8$). 

\smallskip\noindent\textbf{Implementation Details.} We employ standard data augmentations during training, including random cropping and color jitter. Our model is instantiated with the pretrained CLIP-ViT-B~\cite{CLIP} and optimized using Adam~\cite{adam} with cosine learning rate decay. The base learning rate is set to 1e-4 for all datasets. Since most PEFT methods in our comparison have not been evaluated on action recognition, we assess representatives from all four PEFT categories with publicly released code, under a consistent experimental setting. As illustrated in Figure \ref{fig:framework_addtp}, we study two integration schemes for spatial and temporal prompting knowledge: a parallel design and a cascaded design, referred to as \emph{STA-VPT}-Parallel and \emph{STA-VPT}-Cascaded, respectively.

\subsubsection{Comparison with Sequential Prompting Paradigm}
We first compare \emph{STA-VPT} with other sequential prompting methods. To ensure a fair comparison, We approximately equalize the number of tunable parameters across models. The results reported in Table~\ref{tab:table_video} show that our \emph{STA-VPT} with parallel integration scheme achieves the best performance across all datasets, significantly surpassing other visual prompting methods. In particular, \emph{STA-VPT} attains the largest gain over other methods on `SSv2' dataset, where accurate recognition hinges on both robust spatial and temporal modeling, demonstrating the advantages of the spatiotemporally aligned prompting paradigm of our \emph{STA-VPT}. Despite incorporating temporal dynamics explicitly into its visual prompts, \emph{STPN} performs on par with \emph{VPT‑deep}. It is reasonable since we adapt \emph{VPT-deep} to video data by allowing all prompt tokens to interact with all spatiotemporally flattened video tokens across all frames, thereby endowing it with fundamental temporal modeling capability similar to \emph{STPN}.

\subsubsection{Comparison with State-of-the-art Methods} We conduct both quantitative and qualitative evaluations to compare our \emph{STA-VPT} with other state-of-the-art \emph{PEFT} methods.

\smallskip\noindent\textbf{Quantitative Evaluation.}
Table~\ref{tab:table_video} presents a comprehensive comparison of all types of PEFT methods for action recognition, from which we draw three key observations. 1) Our method outperforms both \emph{Full Fine-tune} and \emph{Head Fine-tune} methods significantly. Adapting to video tasks is considerably more challenging than to image tasks due to cross‑modality adaptation and the higher modeling complexity of video data. Our method is particularly superior to `Fine-tune' methods for action recognition in data‑scarce regimes. 2) \emph{AIM} and \emph{ST‑Adapter} are specialized video adapter-tuning methods that incorporate spatiotemporal modeling and exhibit competitive action‑recognition performance. However, our \emph{STA‑VPT‑Parallel} still surpasses them by a large margin, highlighting the advantages of our approach. 3) All video-oriented PEFT methods, including \emph{AIM}, \emph{ST-Adapter} and our \emph{STA-VPT}, substantially outperform image‑oriented PEFT methods on `SSv2' dataset which demands strong spatiotemporal modeling. \emph{STA-VPT-Parallel} achieves the best performance, exceeding the second-best model (\emph{AIM}) by $4\%$.

\begin{table*}[!t]
\centering
\caption{Ablation study on the proposed spatiotemporally aligned prompting paradigm. Parallel spatial-temporal prompting integration is used. }
\label{tab:table_ablation_align}
\renewcommand\arraystretch{1.0} 
\resizebox{0.65\linewidth}{!}{
\begin{tabular}{>{\small}c|>{\small}c|>{\small}c|>{\small}c|>{\small}c|>{\small}c|>{\small}c} 
\toprule
Spatial alignment & Temporal alignment & HMDB51 & SSv2 & UCF101 & K100 & Mean\\
\cmidrule{1-7}
\XSolidBrush &  \XSolidBrush & 71.37 & 39.80 & 94.66 & 85.25 & 72.77\\
\Checkmark &  \XSolidBrush & 72.35 & 43.80 & 94.95 & \textbf{85.70} & 74.20\\
\Checkmark & \Checkmark & \textbf{73.20} & \textbf{49.70} & \textbf{95.48} & \textbf{85.70} & \textbf{76.02}\\
\bottomrule
\end{tabular}
}
\end{table*}

\smallskip\noindent\textbf{Qualitative Evaluation.}
To gain deeper insight into the features learned by STA-VPT and the baselines, we conduct two qualitative studies, as in the previous image prompting experiments. First, we visualize the Grad-CAM~\cite{ref44} feature maps in the last layer of the backbone for these methods, which illustrate the attention regions of
learned features by each method. Figure~\ref{fig:vis_video} presents results for three randomly selected video samples from different datasets. Compared with other methods, our \emph{STA-VPT} exhibits sharper and more accurate focus on action-relevant objects, evidencing its advantage. Furthermore, we visualize t-SNE~\cite{ref43} embeddings of the final-layer backbone features for each method. Figure \ref{fig:video_tsne} shows results on two subsets randomly sampled from HMDB51 and SSv2, respectively. STA-VPT yields more precise, compact class-wise clusters than the baselines, consistent with the quantitative classification results.

\subsubsection{Ablation Study} 
\smallskip\noindent\textbf{Parallel vs Cascaded Spatial-Temporal Prompting Integration.} As reported in Table \ref{tab:table_video}, both STA-VPT variants (parallel and cascaded) exceed other PEFT baselines, with the parallel integration yielding the stronger results. The advantage is especially pronounced on the challenging SSv2 dataset, where both spatial and temporal features are critical for action recognition. We hypothesize that the parallel scheme benefits from fully decoupled prompting without mutual interference. 

\smallskip\noindent\textbf{Effect of SpatioTemporally Aligned Prompting.} To evaluate the spatial and temporal alignments during prompting, the core design of our \emph{STA-VPT}, we conduct ablation study in Table~\ref{tab:table_ablation_align}. The results reveal that both the spatial and temporal alignments are crucial to the performance of action recognition. In particular, each component yields substantial gains on the most challenging dataset, SSv2, validating their effectiveness.



\section{Conclusion}
In this work we have introduced a novel visual prompting paradigm—the SpatioTemporal Aligned Visual Prompt Tuning model (\emph{STA-VPT}). In contrast to the sequential visual prompting paradigm that learns an unordered sequence of prompts, our \emph{STA-VPT} learns a 2D prompt map for images or a 3D prompt volume for videos that is spatially (spatiotemporally) aligned with the input image token map or video token volume. Each prompt token is designated to exclusively prompt knowledge only for the spatiotemporally corresponding visual tokens. This design enables the visual prompts to capture the underlying spatiotemporal structure and facilitates more effective knowledge distillation. Moreover, it allows our \emph{STA-VPT} to conduct fine-grained, individualized prompting with highly granular adaptation to diverse visual patterns, potentially enhancing the prompting performance through fine-grained allocation of prompting capacity. Extensive experiments on both image prompting and video prompting across diverse vision tasks have demonstrated the advantages of our method over other methods for visual prompt tuning.

\bibliographystyle{IEEEtran}
\bibliography{egbib}

\newpage

\vfill

\end{document}